\documentclass{article}

    \PassOptionsToPackage{numbers, sort&compress}{natbib}

    \usepackage[preprint]{neurips_2025}

\usepackage[utf8]{inputenc} 
\usepackage[T1]{fontenc}    
\usepackage{hyperref}       
\usepackage{url}            
\usepackage{booktabs}       
\usepackage{amsfonts}       
\usepackage{graphicx}       
\usepackage{nicefrac}       
\usepackage{microtype}      
\usepackage{xcolor}         
\usepackage{caption}

\usepackage{kotex}
\usepackage{wrapfig}
\usepackage{amsmath}
\usepackage{amssymb}
\usepackage{cleveref}
\usepackage{enumitem}
\usepackage{pifont}
\usepackage{svg}

\crefformat{section}{\S#2#1#3} 
\crefformat{subsection}{\S#2#1#3}
\crefformat{subsubsection}{\S#2#1#3}
\crefformat{appendix}{\S#2#1#3}
\crefformat{figure}{Figure~#2#1#3}

\title{Interpreting Attention Heads for Image-to-Text Information Flow in Large Vision–Language Models}

\author{%
Jinyeong Kim\quad\quad
Seil Kang\quad\quad
Jiwoo Park\quad\quad
Junhyeok Kim\quad\quad
Seong Jae Hwang\thanks{Corresponding author.}\\
Yonsei University\\
\texttt{\{jinyeong1324, seil, wldn1677, timespt, seongjae\}@yonsei.ac.kr}
}

\begin{document}

\maketitle

\vspace{-10pt}
\begin{abstract}
\vspace{-5pt}

Large Vision-Language Models (LVLMs) answer visual questions by transferring information from images to text through a series of attention heads. While this image-to-text information flow is central to visual question answering, its underlying mechanism remains difficult to interpret due to the simultaneous operation of numerous attention heads.
To address this challenge, we propose \textit{head attribution}, a technique inspired by component attribution methods, to identify consistent patterns among attention heads that play a key role in information transfer. Using head attribution, we investigate how LVLMs rely on specific attention heads to identify and answer questions about the main object in an image.
Our analysis reveals that a distinct subset of attention heads facilitates the image-to-text information flow.
Remarkably, we find that the selection of these heads is governed by the semantic content of the input image rather than its visual appearance.
We further examine the flow of information at the token level and discover that (1) text information first propagates to role-related tokens and the final token before receiving image information, and (2) image information is embedded in both object-related and background tokens.
Our work provides evidence that image-to-text information flow follows a structured process, and that analysis at the attention-head level offers a promising direction toward understanding the mechanisms of LVLMs.

\end{abstract}
\section{Introduction}
\label{SECTION:INTRODUCTION}
\vspace{-5pt}

Large Vision-Language Models (LVLMs) have demonstrated remarkable success in a wide range of vision and language tasks. As LVLMs have growing importance in various applications, understanding their inner workings has become increasingly crucial. Uncovering the mechanisms behind LVLMs can enhance transparency~\citep{rauker2023toward}, identify potential biases~\citep{bereska2024mechanistic}, and facilitate the development of more efficient and robust models~\citep{marks2024sparse,he2024matters}, parallel to the recent progress in interpretability research for language models.

However, interpreting LVLMs presents inherent challenges that go beyond those encountered with language-only models. Unlike discrete text tokens in LLMs, which carry well-defined semantic meanings, image tokens in LVLMs act as soft prompts without explicit semantic interpretation. This ambiguity complicates the analysis of internal representations.
Moreover, while most mechanistic interpretability studies in language models rely on short textual inputs~\citep{wang2022interpretability,hanna2023does,lieberum2023does,haklay2025position}, LVLMs process hundreds of image tokens by design. As a result, the mechanisms underlying image-to-text information flow remain largely elusive.

Previous work has primarily focused on the layer level~\citep{neo2024towards,kaduri2024s,jiang2024interpreting}, leaving the role of specific attention heads in facilitating this information transfer unexplored. Identifying such heads could reveal the precise locations where image information is injected and propagated within the model, enabling a more fine-grained understanding of the underlying process.

\begin{figure*}[ht]
    \vskip 0in
    \begin{center}
    \centerline{\includegraphics[width=\textwidth]{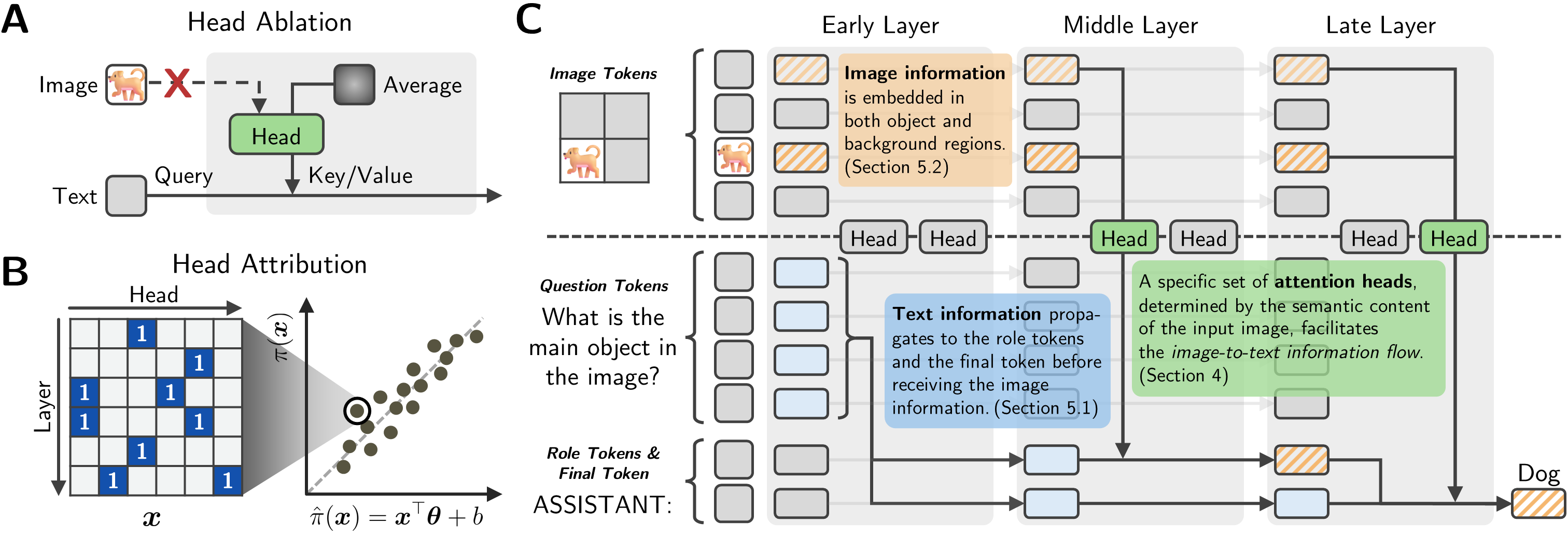}}
    \caption{\textbf{(A)} Illustration of head ablation. To identify the attention heads responsible for the image-to-text information flow, head ablation blocks the image-to-text information flow of an attention head by replacing its image key/value pairs with the baseline representations. Ablating single head cannot identify crucial heads, as the model utilizes multiple heads to process the information.
    \textbf{(B)} Illustration of head attribution. This method estimates each attention head's contribution to the information flow using a linear model that predicts the logit based on which heads are intact or ablated. Through this method, we can identify the pattern of attention heads that are crucial for the image-to-text information flow.
    \textbf{(C)} Overview of the image-to-text information flow. We uncover the mechanism of image-to-text information flow at the levels of attention heads (\cref{SECTION:MAIN_1}) and individual tokens (\cref{SECTION:MAIN_2}).}
    \label{Fig-MAIN}
    \end{center}
    \vskip -0.3in
\end{figure*}

In light of these challenges, this paper conducts a series of experiments to uncover the fine-grained mechanisms underlying visual question answering in LVLMs. Specifically, we seek to identify the critical attention heads responsible for the image-to-text information flow. Once identified, we analyze how these attention heads mediate the transfer of image information by tracing the visual tokens where the information originates and the language tokens that receive it. To achieve this analysis, we focus on the visual object identification task, where the model is asked to identify the main object in a given image.

First, we show that simply ablating single attention heads (\cref{Fig-MAIN}A), as done in LLMs \cite{geva2023dissecting}, is insufficient to identify the key attention heads responsible for image-to-text information flow. This is because the information flow is distributed across multiple heads, making it challenging to isolate a single responsible component.

Therefore, we utilize a systematic approach to quantify the contributions of multiple attention heads to the information flow: \textit{head attribution} (\cref{Fig-MAIN}B), an adapted version of component attribution from \citet{shah2024decomposing}. In this method, we systematically ablate multiple attention heads and use linear regression to estimate each head's contribution to the final logits. The resulting regression model provides a precise estimation of each head's impact on the final logits. Furthermore, since LVLMs may rely on multiple attention heads for image-to-text information flow, head attribution offers a comprehensive explanation of which attention heads are crucial for visual object identification. Our evaluation demonstrates that head attribution outperforms attention-based importance measures and single-head ablations in both faithfulness and completeness.

Head attribution provides a unique lens into the behavior of LVLMs, revealing how they transfer image-to-text information. By analyzing the regression coefficients derived from head attribution, we investigate how attention heads are leveraged for this information flow. Our key findings are as follows. First, attention heads in mid-to-late layers play a pivotal role in transferring image information to text tokens, with a select few heads agonistically utilized across different samples. Second, the contribution of each head is not necessarily correlated with its image attention weights, challenging the common assumption that heads with high attention weights are inherently more important. Lastly, LVLMs employ a similar set of attention heads to process semantically similar objects, suggesting that the model systematically processes objects based on their semantic meanings.

To gain finer-grained insight into the image-to-text information flow, we extend the approach described above to the individual token level. This token-level analysis reveals that the semantics of the question propagate to the role tokens (e.g., ``\textsc{assistant}'') and the final token (``:''), after which image information is transferred to these tokens. We also observe that only a subset of tokens in the object region, along with a few background tokens, contribute to the final prediction. While attention weights correlate with token importance to some extent—explaining the success of recent token reduction methods~\citep{chen2025image,xing2024pyramiddrop}—many high-attention tokens are not necessary for the information flow. This finding suggests that further token reduction in LVLMs could be achieved by quantifying each token’s importance through methods that go beyond attention-based metrics.

In summary, our main contributions are as follows: (1) We provide a holistic understanding of how the image-to-text information flow operates in LVLMs (\cref{Fig-MAIN}C), studied at the levels of attention heads (\cref{SECTION:MAIN_1}) and individual tokens (\cref{SECTION:MAIN_2}). We validate the generalizability of our findings across ten different LVLMs. (2) We demonstrate that head attribution serve as effective methods for explaining the information flow (\cref{SUBSECTION_4_1}, \ref{SUBSECTION_4_2}), offering a novel perspective to identify the mechanisms behind LVLMs. (3) We discuss the broader implications of our findings for mechanistic interpretability and the development of efficient LVLMs (\cref{SECTION:DISCUSSION}).
\section{Related Works}
\label{SECTION:RELATED_WORKS}
\vspace{-5pt}

\textbf{Mechanistic Interpretability.} Mechanistic interpretability (MI) is an emerging research field dedicated to interpret the inner workings of neural networks by reverse-engineering their computations into human-interpretable mechanisms~\citep{olah2022mechanistic}. MI employs various techniques to analyze model internals and understand their decision-making processes~\citep{rai2024practical}. In this section, we briefly review two techniques closely related to our work: causal intervention and component attribution.

\textbf{Causal Intervention.} Causal intervention~\citep{vig2020investigating} treats the neural network as a causal model~\citep{geiger2021causal} and intervenes in the computation of a model component to observe its effect on the prediction. Attention knockout~\citep{geva2023dissecting} can be considered a specialized causal intervention technique for attention mechanisms, where the attention weight from a source of interest to target is set to baseline representations (e.g., zero ablation) to block the information flow.

\textbf{Component Attribution.} Beyond estimating the effect of individual model components through individual ablation, component attribution~\citep{shah2024decomposing} estimates the contributions of multiple components in a more systematic way. Specifically, component attribution ablates subsets of components repeatedly and estimates the contribution of each component via a linear model. We adapt this method for the attention heads of LVLMs, referred to as \textit{head attribution}. Furthermore, while the original work focused on model editing, we repurpose this method to analyze the inner workings of LVLMs by examining the contribution of each head.

\textbf{Interpreting Large Vision-Language Models.} Parallel to the growing interest in interpretability for language models~\citep{ferrando2024primer}, interpreting the inner workings of LVLMs has become a recent focus of research~\citep{dang2024explainable}. In particular, the \textit{image-to-text information flow}, which enables the model to generate text from images, has been widely studied. \cite{jiang2024interpreting} found that image tokens can be projected onto their language vocabulary, and \cite{jiang2024devils} proposed visual information enrichment and semantic refinement as two stages of visual information processing. \cite{neo2024towards} showed that transferring information from object tokens to the final output in mid-to-late layers is necessary for generating text. \cite{kaduri2024s} complemented this finding by revealing that image information is conveyed to the final output through the language context. On the other hand, although not directly focused on interpretability, studies on efficient inference also provide insights into the mechanisms of LVLMs. For example, \cite{chen2025image} and \cite{xing2024pyramiddrop} demonstrated that preserving only image tokens with high attention weights is sufficient for maintaining performance, which suggests that many tokens are redundant in generating outputs. By integrating and clarifying these findings with our experiments, we provide a holistic understanding of image-to-text information flow and its implications for the interpretability and efficiency of LVLMs.
\section{Experimental Setup}
\label{SECTION:SETUP}
\vspace{-5pt}

To systematically examine the mechanism of image-to-text information flow in LVLMs, we introduce a minimal yet natural task, a carefully selected dataset, and a diverse set of models.

\textbf{Task.} This paper seeks to understand the mechanism by which image information is transferred to text tokens and contributes to the final prediction in visual question answering. As a minimal yet natural task within visual question answering, we focus on the \textit{visual object identification task}, in which the model is required to identify the main object in an image. We choose this task because it is a fundamental visual recognition task and is widely used to evaluate the visual capabilities of LVLMs~\citep{rohrbach2018object,li2023evaluating}. Specifically, the task is defined as follows: ``\textsc{user}: ~\textit{$<$image$>$}~ What is the main object in the image? Please answer with a single word. \textsc{assistant}:''.
After the role tokens ``\textsc{assistant}'' and the final token ``:'', the model is expected to generate a concise response (e.g., ``dog'') based solely on the image.
For the ablation studies, we also consider alternative prompts that perform the same task, in order to examine whether LVLMs utilize attention heads based on the general meaning of the task or whether they are sensitive to the specific wording of the prompt. Please refer to \cref{APPENDIX:EXPS} for the alternative prompts we consider.

This experimental design offers several advantages for analysis. First, the task is simple and straightforward, allowing us to focus on the mechanism of image-to-text information flow without the distractions of complex elements such as multi-step reasoning. Second, because the prompt is open-ended and the answer cannot be inferred from it, the model is required to generate its response directly from the image. This minimizes potential biases introduced by the prompt and eliminates the possibility of random guessing~\citep{chen2024we}. Third, we can directly measure the logits due to the single-word answer constraint. Since logits are known to be the most direct indicator of model behavior~\citep{heimersheim2024use}, this design enables a detailed analysis of the model's decision-making process.

\textbf{Dataset.} For image data, we select images containing a single main object from the COCO dataset~\citep{lin2014microsoft} to ensure the task is well-defined and unambiguous. The model is tasked with predicting the object in each image, and we collect 200 samples with correct predictions.
For ablation studies, We also conduct experiments on the DomainNet dataset~\citep{peng2019moment}, which includes diverse images spanning multiple domains (styles). Specifically, we select 300 images from the real, sketch, and clipart domains.

\textbf{Models.} To examine the universality and consistency of our findings across different models, we select ten widely used LVLMs from four families: LLaVA-1.5-7B/13B~\citep{LLaVA1.5}, LLaVA-NeXT-7B/13B~\citep{LLaVANeXT}, InternVL2.5-1B/2B/4B/8B~\citep{InternVL2.5}, and Qwen2-VL-2B/7B~\citep{Qwen2VL}. These models differ in architecture, parameter size, and training scheme, enabling us to assess the generalizability of our findings. In the main paper, we visualize the results from LLaVA-1.5-7B for brevity, with additional results for other models in \cref{APPENDIX:ADDITIONAL}.
\section{Identifying Attention Heads Contributing to Image-to-Text Information Flow}
\label{SECTION:MAIN_1}

In this section, we identify the attention heads that play a crucial role in the image-to-text information flow within LVLMs. We begin by showing that single attention head ablation is insufficient to identify important attention heads. Then, we introduce \textit{head attribution} as a method to measure the contributions of individual attention heads to the final logits (\cref{SUBSECTION_4_1}). We validate the head attribution method by assessing its faithfulness and completeness (\cref{SUBSECTION_4_2}). Finally, we analyze the results of head attribution to gain a deeper insight of how LVLMs leverage attention heads for the transfer of image-to-text information (\cref{SUBSECTION_4_3}).

\subsection{Measuring the Contributions of Attention Heads}
\label{SUBSECTION_4_1}


Image-to-text information flow can be interpreted as a retrieval process, where the text query retrieves image information from the image via the attention mechanism. At each attention head, the text query vector interacts with the key/value matrices of the image tokens to access image information.

\textbf{Attention Knockout (Single Head Ablation).} To determine which attention heads are crucial for image-to-text information flow, we aim to measure the causal effect of each attention head on the final logits by ablating individual heads. Recent LVLM interpretability studies~\citep{neo2024towards, kaduri2024s,serra2024narrow} commonly use attention knockout~\citep{geva2023dissecting} to block the information flow. 
This method adds $-\infty$ before softmax when calculating the attention weights, equivalent to setting the image attention weights and value matrices to zero.
We follow this approach to disable the image-to-text information flow in individual heads.
However, since zero is an arbitrary value that does not reflect the actual distribution of the image tokens, this approach risks introducing undesired out-of-distribution noise~\citep{wang2022interpretability, li2024optimal}.
Instead, we apply mean ablation to block the information flow from the image tokens to the text query. See \cref{APPENDIX:DETAILS} for more details.

\textbf{Single Head Ablation is Insufficient.} We ablate each attention head individually and evaluate the logit difference between the patched and original models. Given the evidence suggesting that attention heads have specialized roles~\citep{zheng2024attention}, we anticipate the logit difference to be significant if an attention head plays an important role in image-to-text information flow. However, the results show that the logit difference is not substantial. For instance, in LLaVA-1.5-7B, no head accounts for more than 5\% of the logit difference in 80\% of the cases (see \cref{APPENDIX:EXPS} for examples).

Based on this observation, we hypothesize that image-to-text information flow is a distributed process involving multiple attention heads, similar to the retrieval process in LLMs~\citep{monea2023glitch}. In this context, quantifying the importance of each attention head by knocking out a single head, which is a common practice in language models~\citep{wang2022interpretability,lieberum2023does}, may be unreliable, as other heads could compensate for the disabled head through self-repair mechanisms~\citep{mcgrath2023hydra,rushing2024explorations}.


\textbf{Method: Head Attribution.} To capture the distributed nature of the image-to-text information flow, we apply head attribution to groups of heads simultaneously rather than isolating individual attention heads, mitigating the interactions between attention heads. After repeatedly collecting logits from the patched models, we leverage component attribution~\citep{shah2024decomposing} to estimate each attention head's contribution to the image-to-text information flow.


Specifically, let 
$\boldsymbol{x} \in \{0,1\}^{LH}$ 
represent a binary vector where each element indicates whether the corresponding attention head is intact ($1$) or ablated ($0$), with $L$ denoting the number of layers and $H$ the number of heads per layer. The final logit, denoted as $\pi(\boldsymbol{x})$, is adjusted by subtracting the average logits over all possible vocabularies~\citep{heimersheim2024use} and is then normalized to the range $[0, 1]$ by setting $\pi(\boldsymbol{0})$ to $0$ and $\pi(\boldsymbol{1})$ to $1$. We estimate the logits by fitting a linear regression model:
\begin{equation}
    \hat{\pi}(\boldsymbol{x}) = \boldsymbol{x} ^\top \boldsymbol{\theta} + b,
\end{equation}
where $\boldsymbol{\theta} \in \mathbb{R}^{LH}$ and $b \in \mathbb{R}$ are the learned parameters. The \textit{attribution coefficients} $\boldsymbol{\theta}$ are interpreted as the contributions of each attention head to the final logits. We refer to this method as \textit{head attribution}. For more technical details, please refer to \cref{APPENDIX:DETAILS}.

\begin{figure}[t!]
    \vskip 0in
    \begin{center}
    \centerline{\includegraphics[width=0.6\columnwidth]{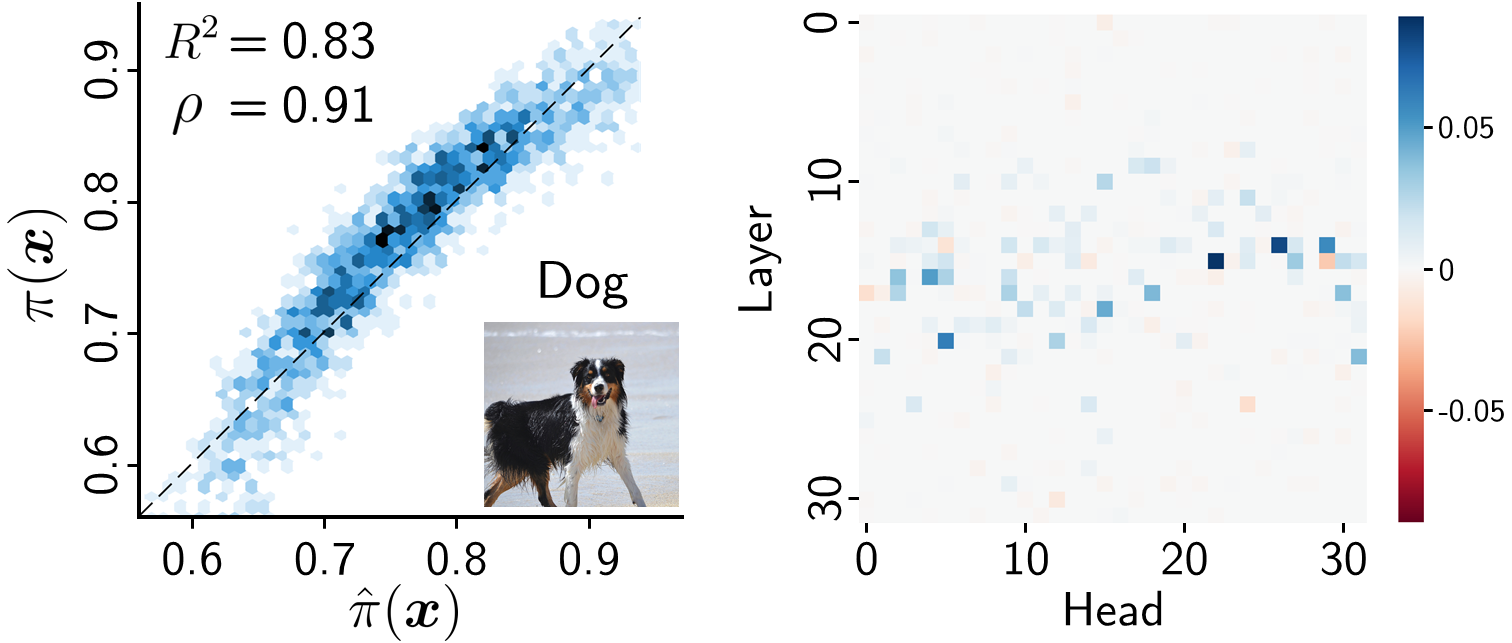}}
    \caption{Example result of head attribution for LLaVA-1.5-7B. \textbf{(Left)} Scatter plot of the ground-truth logit $\pi(\boldsymbol{x})$ and the predicted logit $\hat{\pi}(\boldsymbol{x})$. The contribution of each attention head is well-captured by head attribution.
    \textbf{(Right)} Attribution coefficients $\boldsymbol{\theta}$ for each attention head. Each coefficient represents the contribution of the corresponding head to the prediction.
    }
    \label{Fig-1A}
    \end{center}
    \vspace{-10pt}
\end{figure}

\textbf{Results.} Surprisingly, head attribution can accurately predict the final logit based on which attention heads are intact. \cref{Fig-1A} shows an example of the head attribution results and attribution coefficients $\boldsymbol{\theta}$ for each attention head. The linear model explains the final logit with high accuracy, and the coefficients $\boldsymbol{\theta}$ reveal that multiple attention heads contribute to the image-to-text information flow. Quantitatively, the average explained variance ($R^2$) exceeds $0.77$, and the Pearson correlation coefficient ($\rho$) between $\pi(\boldsymbol{x})$ and $\hat{\pi}(\boldsymbol{x})$ exceeds $0.88$ for all models (see \cref{APPENDIX:ADDITIONAL} for detailed results). These results indicate that the contribution of each attention head is accurately estimated using head attribution.


\subsection{Validating Head Attribution}
\label{SUBSECTION_4_2}

\textbf{Metrics.} If the head attribution method is valid, the coefficients $\boldsymbol{\theta}$ should not only fit well but also reflect the importance of each attention head in the image-to-text information flow. Following \cite{wang2022interpretability, marks2024sparse}, we validate its \textbf{faithfulness} and \textbf{completeness} by evaluating the causal effect. Starting with all heads ablated, we progressively activate the attention heads from the most to the least important based on the coefficients $\boldsymbol{\theta}$. Faithfulness, defined as $\pi(\boldsymbol{x})$, measures the extent to which the selected heads explain the model's actual performance. Completeness, defined as $\pi(\boldsymbol{1} - \boldsymbol{x})$, evaluates whether the selected heads capture all the essential heads. In other words, if other heads can compensate for the selected heads, the method's completeness is compromised. The ideal faithfulness (\textit{i.e.}, the model's actual performance) is $\pi(\boldsymbol{1}) = 1$, and the ideal completeness is $\pi(\boldsymbol{1} - \boldsymbol{1}) = 0$. 




\textbf{Baselines.} We compare the number of heads required to achieve the same level of faithfulness and completeness across three other methods: (1) \textbf{Random}: Heads are selected randomly. (2) \textbf{Attention}: The heads with the highest image attention weights\footnote{To calculate image attention weights, the sum of the image attention weights for each text token is averaged across all text tokens. See \cref{APPENDIX:DETAILS} for more details.} are selected~\citep{zhang2024redundancy,jiang2024devils}, which are frequently presumed to transfer image information to the text query. (3) \textbf{Causal}: The heads with the highest logit difference between the head-ablated and original models are selected (\cref{SUBSECTION_4_1}).
If the head attribution method requires fewer heads to achieve the same level of faithfulness and completeness, it suggests that the method more effectively explains the importance of each attention head. In practice, we measure the minimum number of heads required for faithfulness $> 0.8$ and completeness $< 0.2$ (refer to \cref{APPENDIX:ABLATION} for the ablation study on the threshold).

\textbf{Results.} \cref{Fig-2A}A shows the results of the validation. Head attribution (HeAr) requires the fewest heads to achieve the same faithfulness and completeness compared to other methods. Therefore, we conclude that the head attribution method is valid for measuring the importance of each attention head in image-to-text information flow.

An intriguing finding from the validation is that the number of heads required to achieve the same level of completeness is significantly larger than that required for faithfulness. This trend becomes more pronounced as the model size increases (\cref{Fig-2A}B). Since completeness measures whether the selected heads capture all heads contributing to the final logits, this result suggests that the model may use redundant attention heads~\citep{singh2024needs} to ensure robustness in the image-to-text information flow. Furthermore, as the model size increases, the number of redundant heads also increases, likely due to the growing complexity of the model.

\vspace{-5pt}
\subsection{Interpreting the Behavior of LVLMs with Head Attribution}
\label{SUBSECTION_4_3}

\begin{figure}
    \vskip 0in
    \begin{center}
    \centerline{\includegraphics[width=\textwidth]{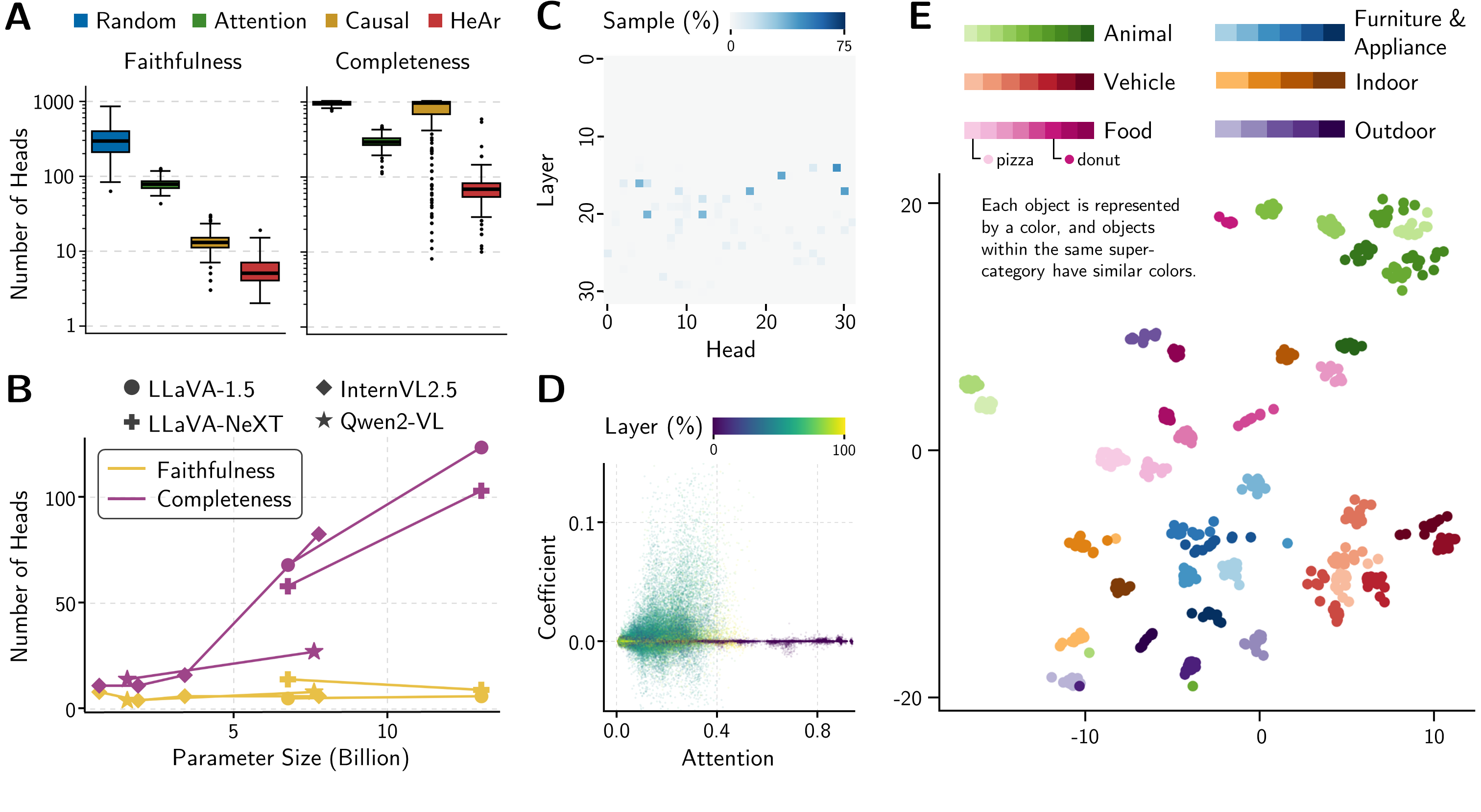}}
    \vspace{-15pt}
    \caption{\textbf{(A)} The minimum number of heads required for faithfulness $> 0.8$ (Left) and completeness $< 0.2$ (Right) for each criterion. Head attribution is abbreviated as HeAr. \textbf{(B)} The trend of the minimum number of heads by head attribution across various models. \textbf{(C)} The portion of samples requiring a given head for faithfulness $> 0.8$. \textbf{(D)} Scatter plot showing the relationship between attribution coefficients $\boldsymbol{\theta}$ and image attention weights. The color indicates relative layer depth. \textbf{(E)} t-SNE visualization of attribution coefficients $\boldsymbol{\theta}$. Each point represents a sample. These plots are for LLaVA-1.5-7B, and the results for other models can be found in \textsection{D}.}
    \label{Fig-2A}
    \end{center}
    \vspace{-24pt}
\end{figure}

In this section, we analyze the head attribution results to gain deeper insights into how LVLMs utilize attention heads for image-to-text information flow. First, we investigate the distribution of important attention heads within the model architecture. Second, we examine whether the importance of the attention heads correlates with the image attention weights to assess whether these weights are reliable indicators of head importance. Finally, we cluster attribution coefficients $\boldsymbol{\theta}$ across different samples to identify common patterns in the attention heads.

\textbf{Distribution of Important Attention Heads.} We visualize the importance of each attention head in the model architecture by counting how many samples rely on a given head to achieve a faithfulness score greater than 0.8 (\cref{Fig-2A}C). Although this varies across models, important heads are generally located in the middle to later layers, in line with the findings of \cite{neo2024towards,jiang2024devils}. Additionally, some heads are consistently important across different samples, while others are only sparsely important, suggesting that LVLMs employ both sample-agnostic and sample-specific attention heads.


\textbf{Relation to Image Attention Weights.} We examine the correlation between attribution coefficients $\boldsymbol{\theta}$ and the image attention weights. As shown in \cref{Fig-2A}D, no significant correlation is observed, indicating that the importance of the attention heads is not directly related to the image attention weights. Therefore, we conclude that image attention weights may not be reliable indicators of the importance of attention heads in image-to-text information flow. We further discuss this in \cref{SECTION:DISCUSSION}. In particular, in the LLaVA family, early-layer heads exhibit high attention weights but are not important. While early-layer heads are known to integrate syntactic information in text~\citep{lad2024remarkable}, we suspect that they are less meaningful in the context of images, where encoding is already effectively handled by the vision encoder.


\textbf{Systematic Patterns of Attention Heads.} We apply t-SNE~\citep{van2008visualizing} to cluster attribution coefficients $\boldsymbol{\theta}$ across different samples (\cref{Fig-2A}E). The clustering pattern reveals that LVLMs tend to utilize similar attention heads when processing the same object, even though the size and position of the object can vary significantly across samples. Moreover, objects with similar semantic meanings (classified by super-categories from COCO~\citep{lin2014microsoft}) are typically clustered together.
Additionally, We apply the logit lens~\citep{logitlens} to the top attention heads ranked by attribution coefficients $\boldsymbol{\theta}$, and visualize the top vocabulary items (see \cref{APPENDIX:EXPS} for examples). We find that some heads are clearly associated with the final answer, indicating that attention heads transfer interpretable semantic information.
Overall, These findings suggest that LVLMs systematically allocate attention heads to process objects based on their semantics.

\textbf{Ablation Studies.}
In LLaVA-1.5-7B, we further conduct two ablation studies to validate the generalizability of our findings. First, we reproduce head attribution using nine additional prompts that perform the visual object identification task, to verify that the observed attention head patterns are not prompt-specific. We find that the cosine similarity between attribution coefficients $\boldsymbol{\theta}$ across different prompts is consistently high (averaging at least 0.91), indicating that the identified patterns are robust to changes in prompt wording.
Second, we apply head attribution to images with diverse visual styles (e.g., real, sketch, and clipart)~\citep{peng2019moment} and observe similarly consistent attention head patterns across styles. These results suggest that the systematic allocation of attention heads is not tied to particular prompts or image styles, but instead reflects a generalizable pattern in LVLMs.
\vspace{-5pt}

\section{Tracing Image-to-Text Information Flow: Token-Level Analysis}
\label{SECTION:MAIN_2}


In \cref{SECTION:MAIN_1}, we identified attention heads that contribute to the image-to-text information flow by applying head attribution. However, this does not provide a complete picture of the information flow. Attention heads transfer information from the source token to the target token. While we know that the source tokens belong to the image and the target tokens belong to the text, the exact positions from which and to which the information flows remain uncertain. To address this, we perform token-level analysis to gain a deeper understanding of the role of these attention heads. Specifically, we identify the text tokens that receive image information (\cref{SUBSECTION_5_1}), and then examine the image tokens that contribute to the information flow (\cref{SUBSECTION_5_2}).

\vspace{-5pt}

\subsection{Which Text Tokens Receive the Image Information?}
\label{SUBSECTION_5_1}


\textbf{Method.} Our text input, ``What is the main object in the image? Please answer with a single word. \textsc{assistant}:'' consists of approximately 20 tokens. Given that information transfer is known to occur only for specific tokens~\citep{wang2022interpretability,jin2024cutting}, we hypothesize that a small number of text tokens play a critical role in this process. To identify the text tokens that receive image information, we measure the causal effect of blocking each text token on the image-to-text information flow through head ablation. A significant drop in the logit indicates that the blocked text token receives image information. Note that we retain only the minimum number of attention heads required for faithfulness $\pi(\boldsymbol{x}^\ast) > 0.8$, which are the most important attention heads for the image-to-text information flow. We then measure the logit difference relative to $\pi(\boldsymbol{x}^\ast)$, as we are interested in the function of attention heads that significantly contribute to the image-to-text information flow.

\begin{figure}
    \vskip 0in
    \begin{center}
    \centerline{\includegraphics[width=\textwidth]{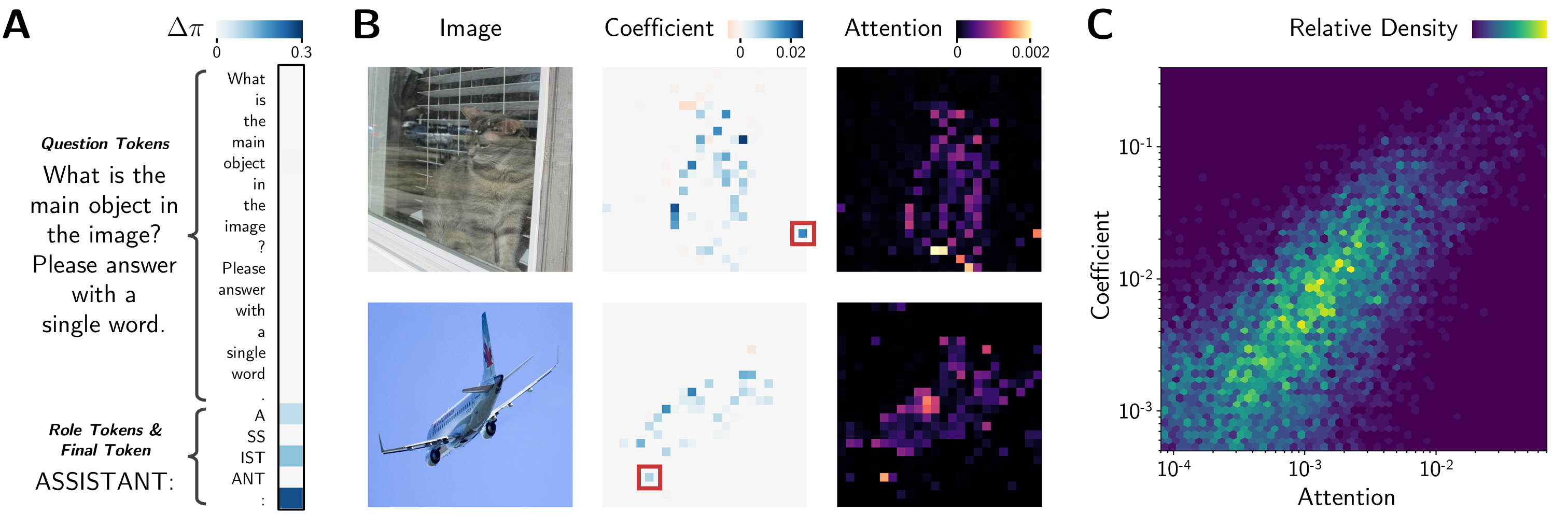}}
    \caption{\textbf{(A)} Logit difference relative to $\pi (\boldsymbol{x}^\ast)$ when blocking each text token. \textbf{(B)} Image, attribution coefficients $\boldsymbol{\theta}$, and image attention weights. Red boxes highlight high-contribution tokens outside the main object region. \textbf{(C)} Scatter plot of attribution coefficients and image attention weights. These plots are for LLaVA-1.5-7B, and the results for other models can be found in \textsection{D}.}
    \label{Fig-4A}
    \end{center}
    \vspace{-20pt}
\end{figure}


\textbf{Results.} As shown in \cref{Fig-4A}A, the logit difference is substantial when we block the role tokens ``\textsc{assistant}'' or the final token before generation ``:''. The causal effect is highest when we block the final token, but the role tokens also have a significant effect. We validate this result by measuring how much of the logit is preserved when we retain only the role tokens and the final token\footnote{More precisely, for example, in LLaVA-1.5-7B, ``\textsc{assistant}:'' is split into five tokens \textsc{a} / \textsc{ss} / \textsc{ist} / \textsc{ant} / :''. Since only the first, third, and final tokens have a high causal effect, we retain only these three tokens.}. In LLaVA-1.5-7B, $97.7 \pm 1.8\%$ of the logit is preserved relative to $\pi(\boldsymbol{x}^\ast)$, indicating that the role tokens and the final token are critical for maintaining the image-to-text information flow. This result clearly explains why blocking only the final token cannot perfectly restrict the information flow~\citep{neo2024towards}, and identifies which text tokens are precisely responsible for the image-to-text information flow, complementing the findings of \cite{kaduri2024s}.


It is quite surprising that no significant information is transferred to the question tokens (``What is the main object in the image?''). Since the question tokens contain the most important semantics for completing the task, it is natural to expect the image information to flow to them preferentially. However, the image information only flows to the role tokens and the final token directly. This result suggests that the model first transfers the semantics of the question to the role tokens and the final token, and then certain attention heads direct the image information to these tokens in response to the question semantics embedded within them.
\vspace{-5pt}
\subsection{Which Image Tokens Contribute to the Flow?}
\label{SUBSECTION_5_2}


\textbf{Method.} After identifying the text tokens that receive the image information, the next step is to determine which image tokens contribute to the information flow. Unlike text tokens, multiple image tokens may contribute, as object information is distributed across several image tokens~\citep{jiang2024interpreting,neo2024towards}. Therefore, simply measuring the causal effect of blocking individual image tokens may not be sufficient to identify the key tokens. Instead, we apply component attribution~\citep{shah2024decomposing} again, this time systematically ablating image tokens rather than attention heads to quantify each image token's contribution to the image-to-text information flow. We focus on LLaVA models for this analysis, as their image encoders preserve the spatial information of the image.


\textbf{Results.} We qualitatively visualize the attribution coefficients $\boldsymbol{\theta}$ for the image tokens in the second column of \cref{Fig-4A}B. By comparing the coefficient with the image, we find that most important image tokens are located in the main object region, which aligns with the findings of \cite{neo2024towards}. However, the coefficient reveals two important insights not addressed in previous studies. First, the model does not utilize all tokens in the main object region; rather, only a sparse subset contributes to the information flow. Second, the model also uses tokens outside the main object region, such as those from the background, to transfer information (highlighted in red boxes in \cref{Fig-4A}B). This seemingly counterintuitive phenomenon may occur because the vision encoder~\citep{CLIP} tends to capture global information in tokens unrelated to the main object~\citep{darcet2023vision,yang2024visionzip}, or because LVLMs treat background tokens as anchor tokens, which language models use to store information~\citep{wang2023label}. A more detailed analysis of this phenomenon is left for future work.


We then compare the attribution coefficients $\boldsymbol{\theta}$ with the attention weights of the image tokens\footnote{We calculate the weighted attention map for each sample. See \cref{APPENDIX:DETAILS} for more details.} (the 3rd column of \cref{Fig-4A}B) to validate whether attention weights can serve as surrogate indicators of the importance of image tokens. In many cases, the attention weights include image tokens with high attribution coefficients. We also visualize the relationship between the attribution coefficients and the attention weights in \cref{Fig-4A}C. Though the trend is not perfect, the attention weights correlate with the attribution coefficients to some extent (Pearson correlation ($\rho$) is $0.70$ for LLaVA-1.5-7B). This result explains why recent token reduction methods based on attention weights~\citep{chen2025image,xing2024pyramiddrop} can be effective.


However, as shown in \cref{Fig-4A}B, many image tokens with high attention weights are not important for the information flow. This result can be explained by (1) attention sink tokens~\citep{kobayashi2020attention,xiao2023efficient}, which are tokens that receive high attention weights but are uninformative, and (2) the fact that some visual tokens contain information but are not useful for predicting the object's name~\citep{jain2019attention,serrano2019attention}. Thus, while retaining tokens with high attention weights could be enough to maintain the information flow, it is essential to be mindful that not all high-attention tokens are necessarily significant.
\section{Conclusion and Discussion}
\label{SECTION:DISCUSSION}
Through \cref{SECTION:MAIN_1}, \ref{SECTION:MAIN_2}, we have shown that head attribution can effectively identify the important heads and tokens for image-to-text information flow. As a result, we now have a holistic understanding of how LVLMs achieve visual object identification task (\cref{Fig-MAIN}C). The process of understanding this simple task provides several insights, including the implications for mechanistic interpretability (MI) and efficient LVLMs. In this section, we discuss these implications and conclude the paper with limitation and future work.


\textbf{Implications for MI.} Interpreting individual attention heads relies on the assumption that some attention heads are monosemantic~\citep{elhage2022toy} (or that the function we aim to understand is localized in a few heads). However, as demonstrated in this work, many attention heads systematically collaborate to process information. This observation aligns with the concept of \textit{attention head superposition}~\citep{jermyn2023attention}, where the model leverages multiple heads, rather than a single head, as an adaptive mechanism for utilizing its limited parameter capacity. In such cases, head attribution offers a promising approach for interpreting distributed mechanisms, as intervening in a single head alone may not fully capture the underlying dynamics.


Especially for interpretability research in LVLMs, our work provides, to the best of our knowledge, the first example of interpretability in individual component-level and pixel-level analysis beyond merely investigating attention weights. Our findings demonstrate that image attention weight is not a reliable indicator of head importance (\cref{SUBSECTION_4_3}), despite its widespread use in prior works~\citep{ilinykh2022attention,zhang2024redundancy,jiang2024devils}. Hence, we encourage the LVLM interpretability community to critically assess the limitations of attention weight-based analyses and to explore alternative approaches for understanding information flow in LVLMs.


\textbf{Implications for Efficient LVLMs.} The computational cost of LVLMs is considerably high due to the large number of image tokens, emphasizing the need for more efficient models. Recent works have introduced strategies to reduce the number of image tokens based on attention weights~\citep{zhang2024sparsevlm,he2024zipvl}. We provide empirical evidence for why these approaches succeed in reducing token count without sacrificing performance, demonstrating a correlation between attention weights and attribution coefficients (\cref{SUBSECTION_5_2}). However, we also show that not all tokens with high attention weights are important. Therefore, we argue that attention-based token selection alone is insufficient, and greater efficiency can be achieved by assessing token importance beyond simple attention weights.

\textbf{Limitation and Future Work.} While our work provides a holistic understanding of image-to-text information flow in LVLMs, several limitations remain. First, our focus is on visual object identification, and the generalization of our findings to other tasks, such as reasoning and localization, is uncertain. A comparative analysis across tasks would deepen our understanding of how LVLMs handle different multimodal tasks. 
Seconds, head attribution requires multiple forward passes, making it computationally expensive. Developing scalable methods for head attribution is an important direction for efficient and practical interpretability~\citep{nanda2023attribution,kramar2024atp}.
\newpage

\bibliography{neurips_2025}





\appendix
\section{Method Details}
\label{APPENDIX:DETAILS}
Through \cref{SECTION:MAIN_1} and \ref{SECTION:MAIN_2}, we have introduced \textit{head attribution} to analyze the information flow in LVLMs. We offer additional details on the background of LVLMs (\cref{APPENDIX:SUBSECTION_1_1}), the methods (\cref{APPENDIX:SUBSECTION_1_2}, \ref{APPENDIX:SUBSECTION_1_3}), and miscellaneous details of the experiments (\cref{APPENDIX:SUBSECTION_1_4}).

\subsection{Background}
\label{APPENDIX:SUBSECTION_1_1}

In this section, we provide a brief overview of the architecture of Large Vision-Language Models (LVLMs) and the attention mechanism employed in these models to contextualize the proposed method.

LVLMs typically process an input image and a text prompt to generate a sequence of text tokens as output. The input image is initially passed through a vision encoder, such as those described in~\cite{CLIP,zhai2023sigmoid}, to produce a visual representation. This representation is then projected into a shared embedding space aligned with the text input. Concurrently, the text input is tokenized and embedded to generate a sequence of text embeddings. Both the visual and text embeddings are fed into a large language model, which is generally implemented as a decoder-only transformer~\citep{vaswani2017attention}.

Let $\boldsymbol{Z} _ {\text{img}} ^ 0 \in \mathbb{R} ^ {P_ {\text{img}} \times d}$ denote the visual embeddings and $\boldsymbol{Z} _ {\text{txt}} ^ 0 \in \mathbb{R} ^ {P_ {\text{txt}} \times d}$ denote the text embeddings, where $P_ {\text{img}}$ and $P_ {\text{txt}}$ represent the number of visual and text embeddings, respectively, and $d$ is the embedding dimension. These embeddings are concatenated as $\boldsymbol{Z} ^ 0 = [\boldsymbol{Z} _ {\text{img}} ^ 0; \boldsymbol{Z} _ {\text{txt}} ^ 0]$ and passed through a decoder-only transformer to generate the output sequence of text tokens. The decoder-only transformer consists of $L$ layers, each comprising a multi-head attention (MHA) mechanism followed by a feed-forward neural network (FFN). For intermediate representations, we define $\boldsymbol{Z} ^ {\ell - 1} = [\boldsymbol{Z} _ {\text{img}} ^ {\ell - 1}; \boldsymbol{Z} _ {\text{txt}} ^ {\ell - 1}]$ as the input to the $\ell$-th layer.

Following the framework proposed by~\cite{elhage2021mathematical}, we adopt the \textit{residual stream} perspective to interpret computations within the transformer. This perspective treats the transformer's computation as a sequence of residual updates, where intermediate representations are iteratively refined by adding the outputs of the MHA and FFN layers to their inputs:
\begin{equation}
\hat{\boldsymbol{Z}} ^ {\ell} = \boldsymbol{Z} ^ {\ell - 1} + \sum _ {h = 1} ^ {H} \text{MHA} ^ {\ell, h} (\boldsymbol{Z} ^ {\ell - 1}), \quad \boldsymbol{Z} ^ {\ell} = \hat{\boldsymbol{Z}} ^ {\ell} + \text{FFN} ^ {\ell} (\hat{\boldsymbol{Z}} ^ {\ell}),
\end{equation}
where $\hat{\boldsymbol{Z}} ^ {\ell}$ represents the output of the MHA layer, $\boldsymbol{Z} ^ {\ell}$ is the output of the FFN layer, and $H$ denotes the number of attention heads per layer. For simplicity, layer normalization~\citep{lei2016layer,zhang2019root} is omitted. Finally, the raw logits are computed via a linear projection of the output of the final layer's text embeddings $\boldsymbol{z} ^ {L} _ {\text{txt}}$ using an unembedding matrix $\boldsymbol{W} _ {U} \in \mathbb{R} ^ {d \times | \mathcal{V} |}$, where $| \mathcal{V} |$ represents the vocabulary size:
\begin{equation}
\text{logits} = \boldsymbol{z} ^ {L} _ {\text{txt}} \boldsymbol{W} _ {U} \in \mathbb{R} ^ {| \mathcal{V} |}.
\end{equation}
We always measure the adjusted logits, obtained by subtracting the average logits across the vocabulary from the raw logits, as recommended by~\cite{heimersheim2024use}.

We focus on the attention mechanism, which facilitates information flow within the model. The MHA mechanism calculates the attention weights $\boldsymbol{A} ^ {\ell, h}$ between the input embeddings $\boldsymbol{Z} ^ {\ell - 1}$ as follows:
\begin{equation} \label{eq:attn_weight}
\boldsymbol{A} ^ {\ell, h} = \text{softmax} \left(\frac{ \boldsymbol{Q} ^ {\ell, h} (\boldsymbol{K} ^ {\ell, h}) ^ \top}{\sqrt{d_k}} + \boldsymbol{M} \right) \in \mathbb{R} ^ {(P_ {\text{img}} + P_ {\text{txt}}) \times (P_ {\text{img}} + P_ {\text{txt}})},
\end{equation}
where $\boldsymbol{Q} ^ {\ell, h} = \boldsymbol{Z} ^ {\ell - 1} \boldsymbol{W} _ {Q} ^ {\ell, h} \in \mathbb{R} ^ {(P_ {\text{img}} + P_ {\text{txt}}) \times d_k}$ and $\boldsymbol{K} ^ {\ell, h} = \boldsymbol{Z} ^ {\ell - 1} \boldsymbol{W} _ {K} ^ {\ell, h} \in \mathbb{R} ^ {(P_ {\text{img}} + P_ {\text{txt}}) \times d_k}$ represent the query and key matrices, respectively. Here, $d_k$ denotes the dimension of the query and key vectors, and $\boldsymbol{M}$ is a mask matrix that prevents attention to future tokens in the sequence, typically by assigning $-\infty$ to the upper triangular part of the matrix. The attention weights are then used to compute the MHA layer's output:
\begin{equation} \label{eq:mha}
\text{MHA}^{\ell, h} (\boldsymbol{Z} ^ {\ell - 1}) = \boldsymbol{A} ^ {\ell, h} \boldsymbol{V} ^ {\ell, h} \boldsymbol{W} _ {O} ^ {\ell, h} \in \mathbb{R} ^ {(P_ {\text{img}} + P_ {\text{txt}}) \times d},
\end{equation}
where $\boldsymbol{V} ^ {\ell, h} = \boldsymbol{Z} ^ {\ell - 1} \boldsymbol{W} _ {V} ^ {\ell, h} \in \mathbb{R} ^ {(P_ {\text{img}} + P_ {\text{txt}}) \times d_k}$ is the value matrix, and $\boldsymbol{W} _ {O} ^ {\ell, h} \in \mathbb{R} ^ {d_k \times d}$ is the output weight matrix.

As discussed in Section 4.1, image-to-text information flow can be understood as a retrieval process. Consider an arbitrary text token $\boldsymbol{z} ^ {\ell - 1} _ {i} \in \mathbb{R} ^ {d}$ at the $\ell$-th layer, which is a row vector from the text matrix $\boldsymbol{Z} ^ {\ell - 1} _ {\text{txt}} \in \mathbb{R} ^ {P_ {\text{txt}} \times d}$. This text token is converted into a query vector $\boldsymbol{q} ^ {\ell, h} _ {i} \in \mathbb{R} ^ {d_k}$ by multiplying it with the query weight matrix $\boldsymbol{W} _ {Q} ^ {\ell, h} \in \mathbb{R} ^ {d \times d_k}$. The query vector $\boldsymbol{q} ^ {\ell, h} _ {i}$ interacts with the key and value matrices of the image tokens $\boldsymbol{K} ^ {\ell, h} _ {\text{img}}$ and $\boldsymbol{V} ^ {\ell, h} _ {\text{img}}$ to retrieve relevant information. Using \Cref{eq:attn_weight}, (\ref{eq:mha}), the attention weights and the MHA output for $\boldsymbol{q} ^ {\ell, h} _ {i}$ are computed as:
\begin{equation} \label{eq:attn_weight_txt}
    \boldsymbol{a} ^ {\ell, h} _ i = \text{softmax} \left(\frac{ \boldsymbol{q} ^ {\ell, h} _ {i} [\boldsymbol{K} ^ {\ell, h} _ {\text{img}}; \boldsymbol{K} ^ {\ell, h} _ {\text{txt}}] ^ \top}{\sqrt{d_k}} + \boldsymbol{m} \right) \in \mathbb{R} ^ {P_ {\text{img}} + P_ {\text{txt}}},
\end{equation}
\begin{equation} \label{eq:attn_weight_txt_2}
    \text{MHA} ^{\ell, h} (\boldsymbol{z} ^ {\ell - 1} _ {i}) = \boldsymbol{a} ^ {\ell, h} _ i[\boldsymbol{V} ^ {\ell, h} _ {\text{img}}; \boldsymbol{V} ^ {\ell, h} _ {\text{txt}}] \boldsymbol{W} _ {O} ^ {\ell, h} \in \mathbb{R} ^ {d}.
\end{equation}
Intuitively, the attention weights $\boldsymbol{a} ^ {\ell, h} _ i$ represent the relevance of other tokens to the text token $\boldsymbol{z} ^ {\ell - 1} _ {i}$, and the MHA layer's output is computed as a weighted sum of these tokens, with the attention weights serving as the coefficients.

\subsection{Head Ablation}
\label{APPENDIX:SUBSECTION_1_2}

In this section, we formally explain the concept of head ablation utilized in \cref{SUBSECTION_4_1}.

Attention knockout, or head ablation~\citep{geva2023dissecting} is a widely used method for analyzing information flow in attention mechanisms. In the context of LVLMs, it is specifically employed to measure image-to-text information flow~\citep{neo2024towards,kaduri2024s,serra2024narrow}. The method involves setting the attention weights between image and text tokens to baseline values and observing the resulting effect on the model's performance. Head ablation replaces the image key and value matrices with averaged values from other images.




Specifically, we collect the key and value matrices of image tokens from 100 images in the COCO~\citep{lin2014microsoft} dataset and compute the average key matrices $\boldsymbol{K} ^{\ell, h} _ {\text{img, avg}}$ and value matrices $\boldsymbol{V} ^{\ell, h} _ {\text{img, avg}}$. The image key and value matrices in the attention mechanism are then replaced as follows:
\begin{equation}
    \boldsymbol{K} ^{\ell, h} _ {\text{img}} \leftarrow \boldsymbol{K} ^{\ell, h} _ {\text{img, avg}}, \quad \boldsymbol{V} ^{\ell, h} _ {\text{img}} \leftarrow \boldsymbol{V} ^{\ell, h} _ {\text{img, avg}}.
\end{equation}
Finally, the attention weights and the MHA layer's output are recomputed using \Cref{eq:attn_weight_txt}, (\ref{eq:attn_weight_txt_2}). By substituting the image key and value matrices with their average values, head ablation effectively eliminates information flow while preserving the natural attention distribution of the model.


\subsection{Head Attribution}
\label{APPENDIX:SUBSECTION_1_3}

In this section, we provide more details on the \textit{head attribution} method introduced in \cref{SUBSECTION_4_1}.

Let $\boldsymbol{x} \in \{ 0, 1 \}^{LH}$ denote a binary vector, where $L$ is the number of layers and $H$ is the number of attention heads per layer. An element $x_n$ of $\boldsymbol{x}$ is set to $1$ if the $n$-th attention head\footnote{The attention heads are indexed in row-major order; that is, the first $H$ elements correspond to the attention heads of the first layer, the next $H$ elements correspond to those of the second layer, and so on. In other words, $h$-th attention head of the $\ell$-th layer is indexed as $n = \ell \times H + h$, where $\ell \in \{0, 1, \ldots, L - 1\}$ and $h \in \{0, 1, \ldots, H - 1\}$.} is intact, and $0$ if it is ablated. For each sample, we ablate $p = 75\%$ of the attention heads in the model\footnote{See \cref{APPENDIX:SUBSECTION_2_2} for the ablation study on the ablation ratio $p$.} $M = 10,000$ times to obtain $\mathcal{D} = \{ (\boldsymbol{x}^{(1)}, \pi(\boldsymbol{x}^{(1)})), \ldots, (\boldsymbol{x}^{(M)}, \pi(\boldsymbol{x}^{(M)})) \}$, where $\pi(\boldsymbol{x})$ denotes the normalized logit as described in \cref{SUBSECTION_4_1}. The obtained dataset $\mathcal{D}$ is then split into two subsets: $\mathcal{D}_{\text{train}}$ and $\mathcal{D}_{\text{test}}$, containing $80\%$ and $20\%$ of the samples, respectively.

We train a linear regression model $\hat{\pi}(\boldsymbol{x}) = \boldsymbol{x} ^ \top \boldsymbol{\theta} + b$ on $\mathcal{D}_{\text{train}}$ to predict the normalized logit $\pi(\boldsymbol{x})$ from the binary vector $\boldsymbol{x}$. The model is trained with elastic net regularization~\citep{Zou_Hastie_2005}, implemented using Scikit-learn~\citep{scikit-learn}. Specifically, our hyperparameters are set as follows:
\begin{equation}
    \texttt{ElasticNet(alpha = 0.0005, l1\_ratio = 0.5, max\_iter = 1000)}.
\end{equation}
The result of linear regression is evaluated on $\mathcal{D}_{\text{test}}$ using the explained variance ($R^2$) and the Pearson correlation coefficient ($\rho$) between the predicted logits $\hat{\pi}(\boldsymbol{x})$ and the true logits $\pi(\boldsymbol{x})$, following \cite{shah2024decomposing}.

\subsection{Miscellaneous Details}
\label{APPENDIX:SUBSECTION_1_4}

To enhance reproducibility and clarity, we provide further details on the experiments described in \cref{SECTION:MAIN_1} and \ref{SECTION:MAIN_2}.

\textbf{Model Details on Head Ablation Applied to Individual Attention Heads (\cref{SUBSECTION_4_1}).} Let $\boldsymbol{x}_{\neg{n}}$ denote a binary vector in which the $n$-th attention head is ablated (\textit{i.e.}, $x_n = 0$, while all other elements of $\boldsymbol{x}_{\neg{n}}$ are set to $1$). The logit difference is then computed as $\Delta \pi_n = 1 - \pi(\boldsymbol{x}_{\neg{n}})$.

\textbf{Image Attention Weights for Each Attention Head (\cref{SUBSECTION_4_2} and \ref{SUBSECTION_4_3}).} We calculate image attention weights for each attention head and use them as a criterion to select the most important attention heads in \cref{Fig-2A}A. These weights are then compared with attribution coefficients in \cref{Fig-2A}D. The image attention weights are determined by averaging the attention weights of image tokens across text tokens. Specifically, given the attention weights $\boldsymbol{A} ^ {\ell, h} \in \mathbb{R} ^ {(P_ {\text{img}} + P_ {\text{txt}}) \times (P_ {\text{img}} + P_ {\text{txt}})}$ from \Cref{eq:attn_weight}, the image attention weight for each head is computed as follows:
\begin{equation}
\text{Image Attention Weight} = \frac{\sum _ {i = P_ {\text{img}} + 1} ^ {P_ {\text{img}} + P_ {\text{txt}}} \sum _ {j = 1} ^ {P_ {\text{img}}} a ^ {\ell, h} _ {i, j}}{P_ {\text{txt}}},
\end{equation}
where $a ^ {\ell, h} _ {i, j}$ represents the element of $\boldsymbol{A} ^ {\ell, h}$ at the $i$-th row and $j$-th column. Similar methods for calculating attention weights per head or per layer have been employed in various studies~\citep{ilinykh2022attention,zhang2024redundancy,kaduri2024s,jiang2024devils,serra2024narrow,wu2024accelerating,ye2024fit}.

\textbf{More Details on t-SNE Visualization (\cref{SUBSECTION_4_3}).} For the t-SNE visualization in \cref{Fig-2A}E and \ref{Fig-XD4}, we collect additional samples from the COCO dataset to enhance the diversity of images for each object. Specifically, objects are included if LLaVA-1.5-7B generates at least 10 correct samples for them. From these, we randomly select up to 15 samples per object and repeat the head attribution process to compute attribution coefficients $\boldsymbol{\theta}$.

\textbf{More Details on Text Token Experiments (\cref{SUBSECTION_5_1}).} The experiments described in \cref{SECTION:MAIN_2} retain only the minimum number of attention heads required to maintain faithfulness above 0.8. This approach focuses on understanding the behavior of the most critical attention heads for image-to-text information flow. To formalize this, we define a binary vector $\boldsymbol{x}^\ast$, where each element is set to $1$ if the corresponding attention head is essential to ensure faithfulness $> 0.8$, and $0$ otherwise.

Since head ablation can be selectively applied to subsets of text or image tokens as explained in \cref{APPENDIX:SUBSECTION_1_2}, we introduce binary vectors $\boldsymbol{u} \in \{ 0, 1 \}^{P_{\text{txt}}}$ and $\boldsymbol{v} \in \{ 0, 1 \}^{P_{\text{img}}}$, where $P_{\text{txt}}$ and $P_{\text{img}}$ represent the total number of text and image tokens, respectively. Each element $u_i$ of $\boldsymbol{u}$ is set to $1$ if the $i$-th text token remains intact, and $0$ if it is ablated. Similarly, each element $v_j$ of $\boldsymbol{v}$ is set to $1$ if the $j$-th image token remains intact, and $0$ if it is ablated.

The normalized logit, computed using only the attention heads required for faithfulness $> 0.8$ (denoted as $\boldsymbol{x}^\ast$), with intact text queries and image tokens, is denoted as $\pi(\boldsymbol{x}^\ast; \boldsymbol{u}, \boldsymbol{v})$. When all text and image tokens are intact, $\pi(\boldsymbol{x}^\ast; \boldsymbol{1}, \boldsymbol{1})$ simplifies to $\pi(\boldsymbol{x}^\ast)$, as defined in \cref{SECTION:MAIN_1}.

In \cref{SUBSECTION_5_1}, we aim to measure the causal effect of blocking image information flow to individual text tokens. To achieve this, we define $\boldsymbol{u}_{\neg{i}}$ as a binary vector where the $i$-th text token is ablated (i.e., only $u_i = 0$, while all other elements of $\boldsymbol{u}_{\neg{i}}$ remain $1$). The logit difference relative to $\pi(\boldsymbol{x}^\ast)$ is then computed as:
\begin{equation} \label{eq:logit_diff}
    \Delta \pi_i = 1 - \frac{\pi(\boldsymbol{x}^\ast; \boldsymbol{u}_ {\neg{i}}, \boldsymbol{1})} {\pi(\boldsymbol{x}^\ast)}.
\end{equation}
If $\Delta \pi_i > 0.05$, we conclude that the $i$-th text token is important for image-to-text information flow.

Once the important text tokens are identified, we further evaluate whether they alone are sufficient for recovering the information flow. To do this, we define a binary vector $\boldsymbol{u}^\ast$, where $u_i^\ast = 1$ if $\Delta \pi_i > 0.05$ and $u_i^\ast = 0$ otherwise. The corresponding logit relative to $\pi(\boldsymbol{x}^\ast)$, computed as $\pi(\boldsymbol{x}^\ast; \boldsymbol{u}^\ast, \boldsymbol{1}) / \pi(\boldsymbol{x}^\ast)$, indicates the extent to which the important text tokens restore the information flow. If the logit is sufficiently high, we conclude that the identified important text tokens are adequate for preserving image-to-text information flow.

\textbf{More Details on Image Token Experiments (\cref{SUBSECTION_5_2}).} In \cref{SUBSECTION_5_2}, we extend the head attribution method to image tokens. Following a similar approach to \cref{APPENDIX:SUBSECTION_1_3}, we ablate $p = 75\%$ of the image tokens in the model $M = 10,000$ times to generate the dataset $\mathcal{D} = \{ (\boldsymbol{v}^{(1)}, \tilde{\pi}(\boldsymbol{x}^\ast; \boldsymbol{u}^\ast, \boldsymbol{v}^{(1)})), \ldots, (\boldsymbol{v}^{(M)}, \tilde{\pi}(\boldsymbol{x}^\ast; \boldsymbol{u}^\ast, \boldsymbol{v}^{(M)})) \}$, where $\tilde{\pi}(\boldsymbol{x}^\ast; \boldsymbol{u}^\ast, \boldsymbol{v})$ denotes the normalized logit computed as:
\begin{equation}
    \tilde{\pi}(\boldsymbol{x}^\ast; \boldsymbol{u}^\ast, \boldsymbol{v}) = \frac{\pi(\boldsymbol{x}^\ast; \boldsymbol{u}^\ast, \boldsymbol{v})}{\pi(\boldsymbol{x}^\ast; \boldsymbol{u}^\ast, \boldsymbol{1})}.
\end{equation}
We then train a linear regression model, $\hat{\pi}(\boldsymbol{x}^\ast; \boldsymbol{u}^\ast, \boldsymbol{v}) = \boldsymbol{v} ^ \top \boldsymbol{\theta} + b$, to predict $\pi(\boldsymbol{x}^\ast; \boldsymbol{u}^\ast, \boldsymbol{v})$ from the binary vector $\boldsymbol{v}$. The model is trained using the same hyperparameters as described in \cref{APPENDIX:SUBSECTION_1_3}.

\textbf{Weighted Image Attention Weights for Each Sample (\cref{SUBSECTION_5_2}).} Since image attention weights are computed for each attention head, we calculate the weighted image attention weights as a representative value for each sample. Let $\boldsymbol{A} ^ n = \boldsymbol{A} ^ {\ell, h}$, where $n = \ell \times H + h$. The weighted image attention weights $\boldsymbol{w} \in \mathbb{R} ^ {P_ {\text{img}}}$ are defined as:
\begin{equation}
\boldsymbol{w} = \sum_ {n=0} ^ {LH-1} \left[ (x_n ^\ast \cdot \theta_n) \sum _ {i = P_ {\text{img}} + 1} ^ {P_ {\text{img}} + P_ {\text{txt}}} \left\{ (u_i ^\ast \cdot \Delta \pi_i) \cdot \boldsymbol{a} ^ n _ i [\colon P _ {\text{img}}] \right\} \right],
\end{equation}
where $\boldsymbol{a} ^ n _ i$ represents the $i$-th row of $\boldsymbol{A} ^ n$,
$\theta_n$ is the $n$-th element of the attribution coefficients $\boldsymbol{\theta}$ derived from head attribution,
$x_n ^\ast$ is the $n$-th element of $\boldsymbol{x}^\ast$,
$u_i ^\ast$ is the $i$-th element of $\boldsymbol{u}^\ast$, and
$\Delta \pi_i$ is the logit difference for the $i$-th text token (\Cref{eq:logit_diff}).
Intuitively, the weighted image attention weights $\boldsymbol{w}$ consider the importance of each attention head and text token in determining the attention weights for image tokens.
The weighted image attention weights $\boldsymbol{w}$ are reshaped into dimensions $\sqrt{P_ {\text{img}}} \times \sqrt{P_ {\text{img}}}$ to visualize the attention distribution in \cref{Fig-4A}B and to create joint distribution plots in \cref{Fig-4A}C.

\textbf{Computational Resources.} All experiments were conducted on  a single NVIDIA RTX 6000 GPU. For LLaVA-1.5-7B, processing a single input example requires approximately 0.81 GPU hours. The total GPU hours for a complete experiment can be calculated by multiplying this value by the number of images. For example, in the case of COCO, using 200 images results in approximately 160 GPU hours.

\section{Ablation Studies}
\label{APPENDIX:ABLATION}
\begin{figure}
    \begin{center}
    \centerline{\includegraphics[width=\columnwidth]{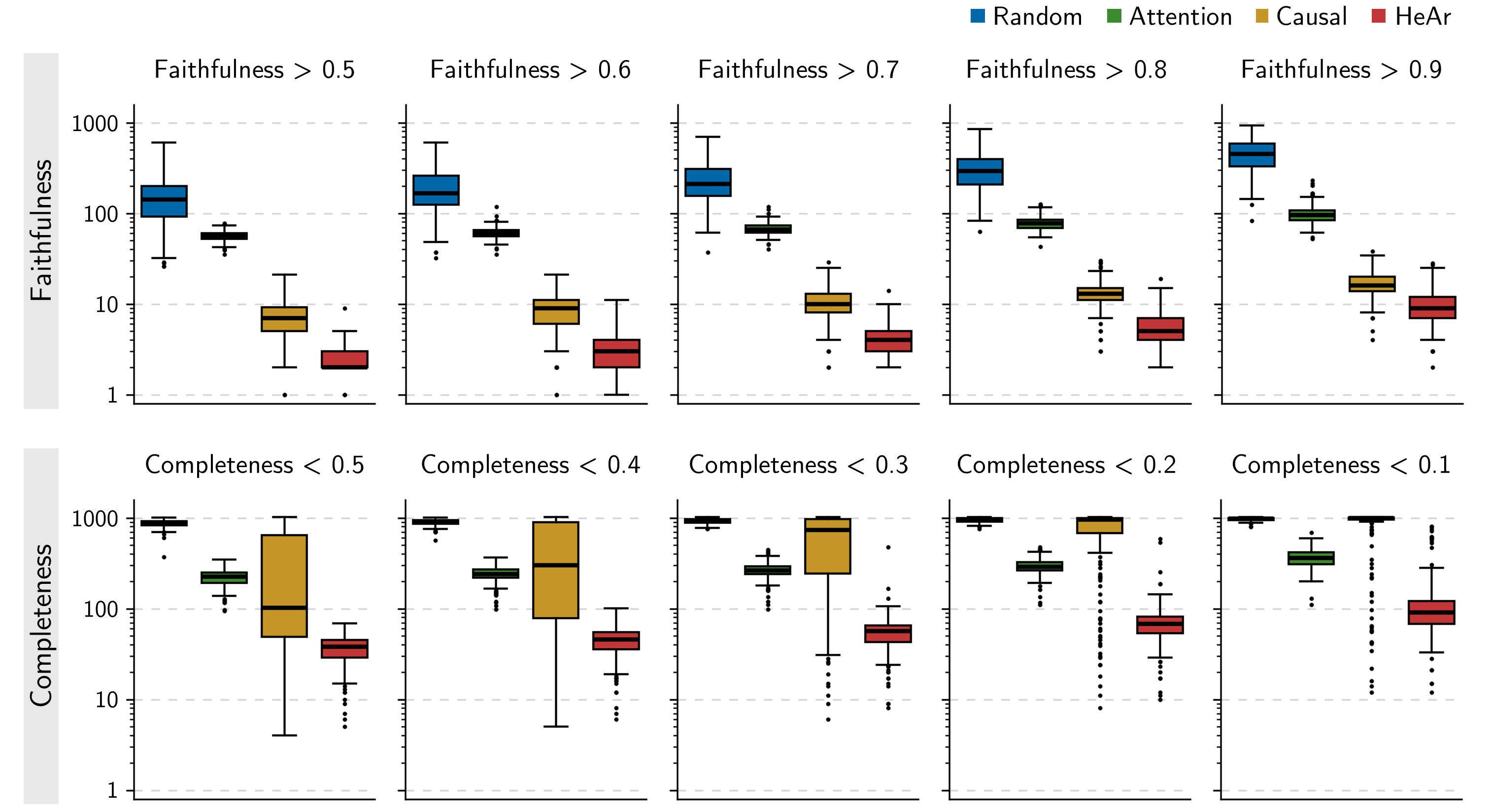}}
    \vskip -0.05in
    \caption{The minimum number of heads required for each threshold of faithfulness (Top) and completeness (Bottom) in LLaVA-1.5-7B. Related to \cref{Fig-2A}A.}
    \label{Fig-XB1}
    \end{center}
    \vskip -0.2in
\end{figure}
\begin{figure}
    \vskip 0in
    \begin{center}
    \centerline{\includegraphics[width=0.85\columnwidth]{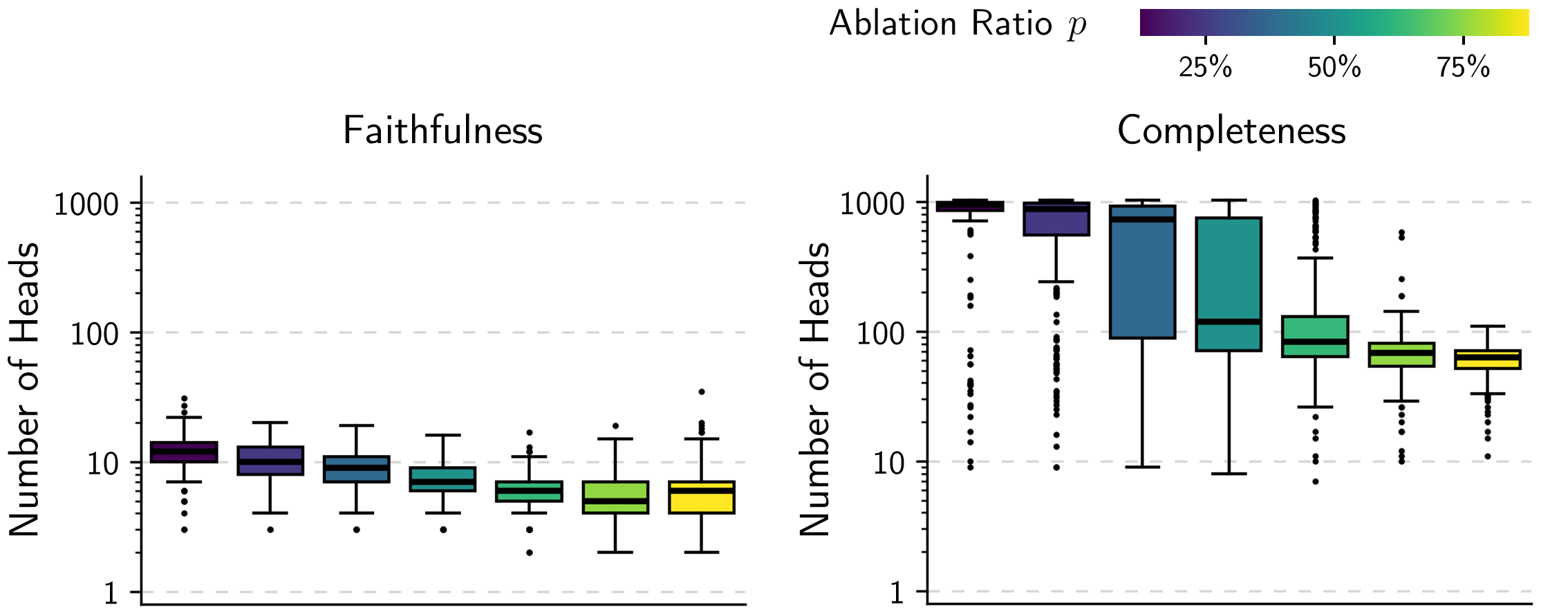}}
    \caption{The minimum number of heads for faithfulness $> 0.8$ (Left) and completeness $< 0.2$ (Right) for each ablation ratio $p$ in LLaVA-1.5-7B. Related to \cref{Fig-2A}A.}
    \label{Fig-XB2}
    \end{center}
    \vskip -0.2in
\end{figure}

\subsection{Ablation Study on Threshold for Validating Head Attribution}
\label{APPENDIX:SUBSECTION_2_1}

In \cref{SUBSECTION_4_2}, we validate head attribution by counting the number of heads required to achieve faithfulness $> 0.8$ and completeness $< 0.2$. In this section, we conduct an ablation study to examine how different thresholds affect the number of heads needed to achieve various levels of faithfulness and completeness. Specifically, we vary the faithfulness threshold from $0.5$ to $0.9$ and the completeness threshold from $0.5$ to $0.1$.

We present the number of heads required to achieve these faithfulness and completeness levels in \cref{Fig-XB1}. As the faithfulness threshold increases and the completeness threshold decreases, the number of required heads increases. Nevertheless, across all thresholds, the head attribution method consistently requires fewer heads to attain the same faithfulness and completeness levels.

\subsection{Ablation Study on Ablation Ratio of Head Attribution}
\label{APPENDIX:SUBSECTION_2_2}

In the main paper, we set the ablation ratio $p$ to $75\%$ for the head attribution method. In this section, we conduct an ablation study to examine how different ablation ratios affect the results of the head attribution method. Specifically, we vary the ablation ratio from $12.5\%$ to $87.5\%$.

As shown in \cref{Fig-XB2}, sufficiently high ablation ratios ($> 50\%$) require fewer heads to achieve the same level of faithfulness and completeness. This result aligns with the motivation behind the head attribution method, which mitigates interactions between heads by ablating multiple heads simultaneously. Additionally, the results remain consistent across sufficiently high ablation ratios, indicating that the head attribution method operates robustly within a reasonable range of ablation ratios.
\section{Additional Experiments}
\label{APPENDIX:EXPS}
\begin{figure}
    \begin{center}
    \centerline{\includegraphics[width=\columnwidth]{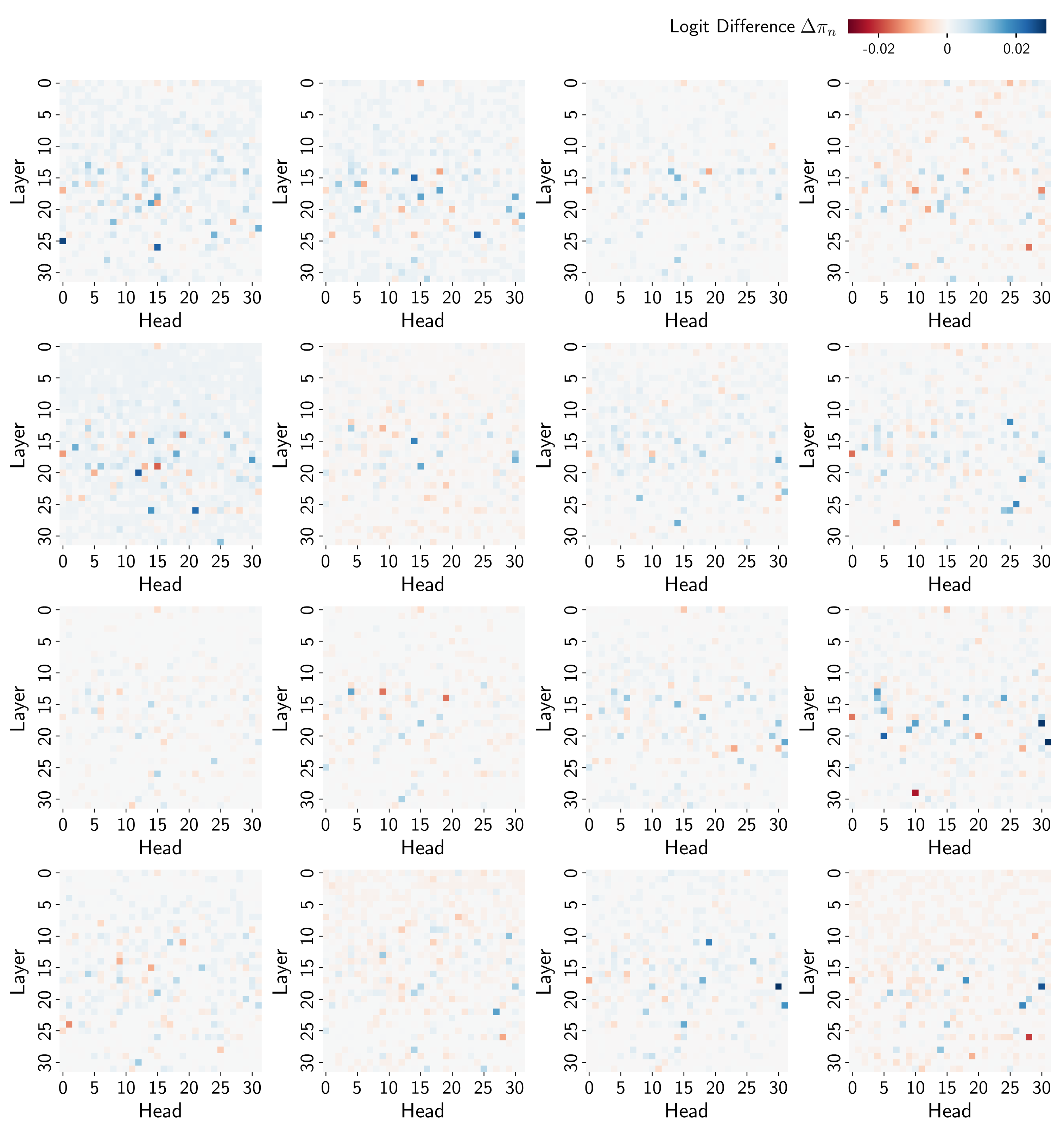}}
    \caption{Head ablation results for individual attention heads in LLaVA-1.5-7B. Each subfigure presents the results of individual head ablation for a randomly selected image sample. The logit difference $\Delta \pi_n$ is not significant for most samples, suggesting that the image-to-text information flow is distributed across multiple attention heads and individual head ablation does not provide a consistent explanation for the model’s behavior (\Cref{APPENDIX:SUBSECTION_3_1}). Related to \cref{SUBSECTION_4_1}.}
    \label{Fig-XC1}
    \end{center}
\end{figure}

\subsection{Head Ablation on Individual Attention Heads}
\label{APPENDIX:SUBSECTION_3_1}

As described in \cref{SUBSECTION_4_1}, we apply head ablation to each attention head individually. We then measure the logit difference $\Delta \pi_n$ between the original and patched models for each head $n$ (see \cref{APPENDIX:SUBSECTION_1_4} for technical details). As shown in \cref{Fig-XC1}, the logit difference is not significant for most samples. These results suggest that image-to-text information flow is distributed across many attention heads rather than being concentrated in a single specialized head. Therefore, we introduce head attribution to more effectively measure each head’s contribution to image-to-text information flow.

\subsection{Logit Lens Analysis}
\label{APPENDIX:SUBSECTION_3_2}

\begin{figure}
    \vskip 0in
    \begin{center}
    \centerline{\includegraphics[width=1\columnwidth]{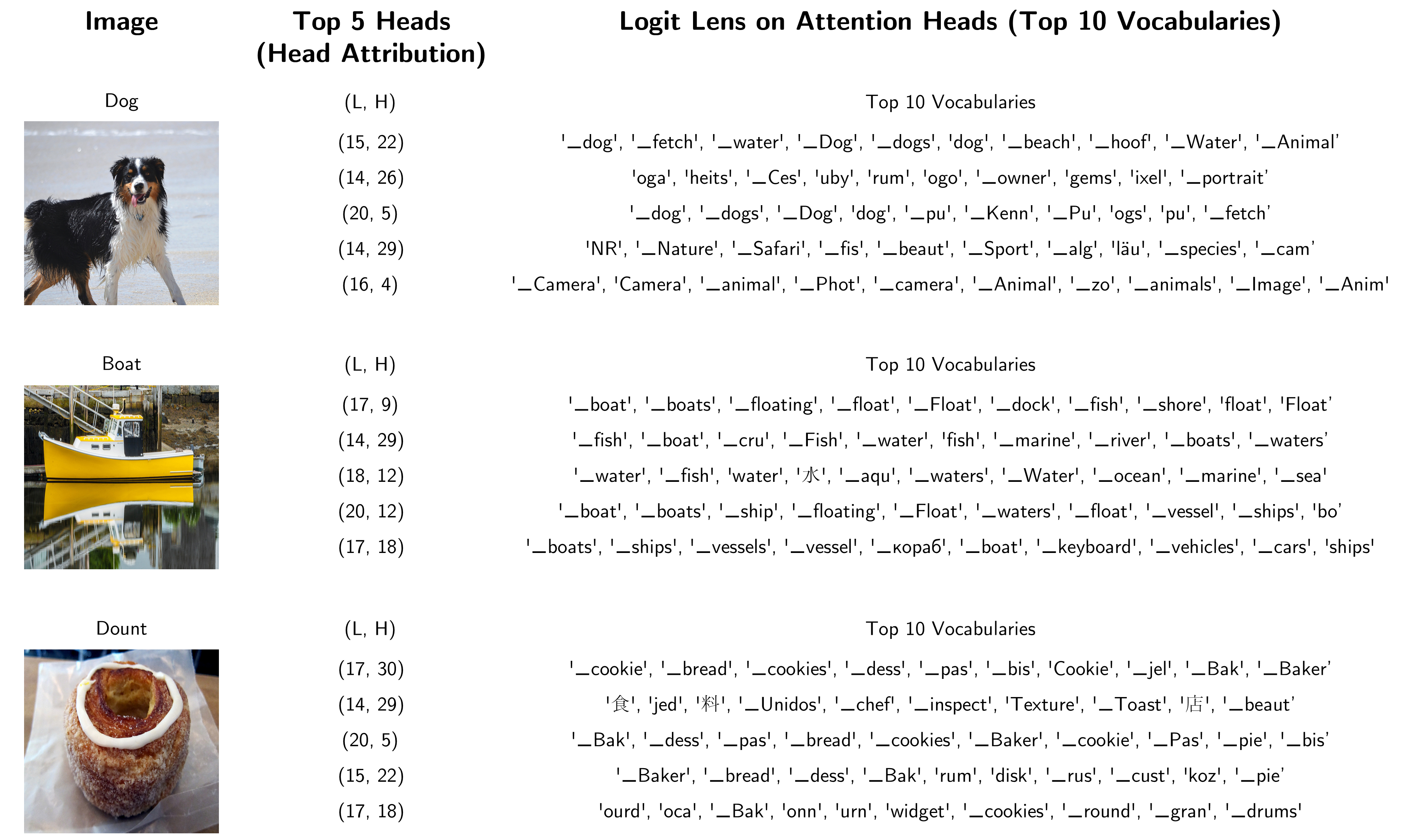}}
    \caption{Logit lens analysis on top-ranked attention heads, selected by attribution coefficients $\boldsymbol{\theta}$. Top-ranked attention heads often transfer semantically meaningful information aligned with the final textual output (\Cref{APPENDIX:SUBSECTION_3_2}).}
    \label{Fig-Rebuttal-DLA}
    \end{center}
    \vskip -0.2in
\end{figure}

To further understand the role of attention heads in image-to-text information transfer, we apply logit lens~\citep{logitlens} to the top 5 attention heads ranked by attribution coefficients $\boldsymbol{\theta}$ and visualize the top 10 vocabulary items in \Cref{Fig-Rebuttal-DLA}.
We find that some heads are clearly associated with the final answer, while others are only weakly related or primarily encode background information.
Although certain influential heads are not interpretable in the vocabulary space, the logit lens reveals that most top-ranked heads transfer semantically relevant information in text, rather than contributing in an uninterpretable way.
This suggests that some degree of image-to-text transformation may have already occurred at the image token level, and the attention heads retrieve this partially converted information.

\subsection{Generalization of Attribution Coefficients Across Prompts}
\label{APPENDIX:SUBSECTION_3_3}

In the main paper, we fix the prompt, ``What is the main object in the image? Please answer with a single word.'' and measure the attribution coefficients $\boldsymbol{\theta}$ for each attention head. We also examine whether LVLMs utilize attention heads based on the meaning of the task or if they are sensitive to the wording of the prompt. To investigate this, we assess whether the attribution coefficients, which represent the contribution of each attention head to the image-to-text information flow, remain consistent across different prompts. If the attribution coefficients are stable across varying prompts, it suggests that LVLMs systematically rely on attention heads according to the task’s meaning rather than the specific wording of the prompt.

We ask ChatGPT to generate nine different prompts, each with different wording but maintaining the same visual object identification task. The prompts are as follows:
\begin{enumerate}
    \item Identify the primary subject in this picture with a single word.
    \item Please specify the key item in this image using one word only.
    \item Please give me the main subject in this picture, stated in a single word.
    \item Provide a one-word term for the principal object depicted here.
    \item In one word, what is the primary focus of the photo?
    \item Name the primary element in the photo, limiting your answer to one word.
    \item Which single word best describes the main subject of this picture?
    \item Using a single word, name the object that stands out the most in this image.
    \item Offer the primary object's name from the picture, restricted to a single word.
\end{enumerate}
We collect 40 samples for each prompt and measure the cosine similarity between the attribution coefficients obtained from the original prompt and those from each of the nine new prompts.

\begin{table}
\caption{Cosine similarity of the attribution coefficients $\boldsymbol{\theta}$ across different prompts. The consistency of the attribution coefficients across prompts suggests that LVLMs utilize attention heads based on the task rather than the specific wording of the prompt (\Cref{APPENDIX:SUBSECTION_3_3}).}
\label{Table-XC1}
\begin{center}
\begin{small}
\resizebox{\columnwidth}{!}{%
\begin{tabular}{lccccccccc}
\toprule
Prompts & Prompt 1 & Prompt 2 & Prompt 3 & Prompt 4 & Prompt 5 & Prompt 6 & Prompt 7 & Prompt 8 & Prompt 9 \\
\midrule
Similarity & $0.97 \pm 0.02$ & $0.96 \pm 0.03$ & $0.97 \pm 0.02$ & $0.96 \pm 0.02$ & $0.91 \pm 0.03$ & $0.94 \pm 0.02$ & $0.94 \pm 0.03$ & $0.93 \pm 0.04$ & $0.94 \pm 0.03$ \\
\bottomrule
\end{tabular}%
}
\end{small}
\end{center}

\end{table}

As shown in \Cref{Table-XC1}, the cosine similarity between the attribution coefficients from the original prompt and those from the nine new prompts is high, with an average similarity of 0.95. These results suggest that the attribution coefficients generalize across different prompts, indicating that LVLMs systematically utilize attention heads based on the task's meaning rather than the precise wording of the prompt.

\subsection{Generalization of Attribution Coefficients Across Image Styles}
\label{APPENDIX:SUBSECTION_3_4}

\begin{figure}
    \vskip 0in
    \begin{center}
    \centerline{\includegraphics[width=1\columnwidth]{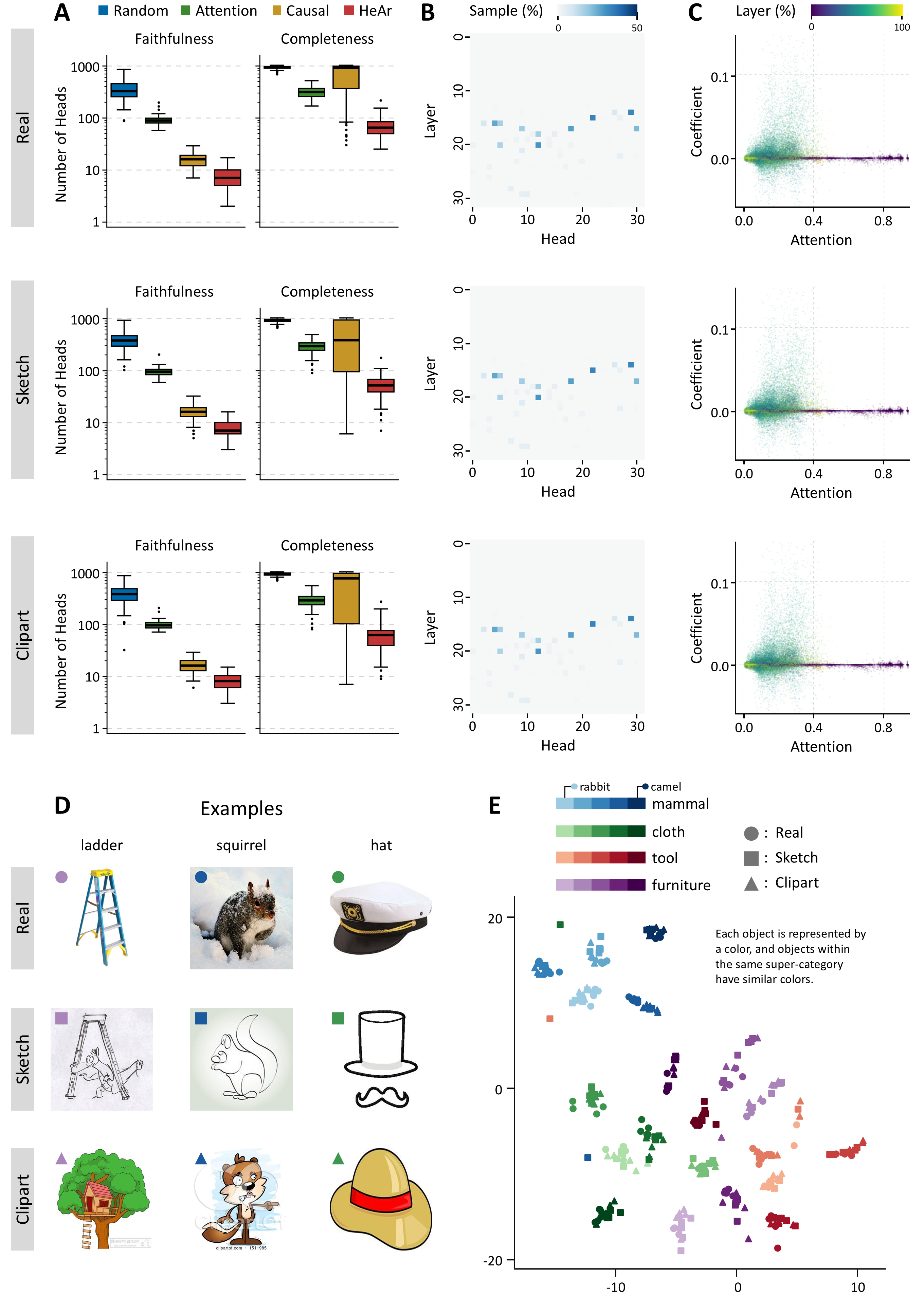}}
    \caption{DomainNet experiment. Despite large variations in image style, attention head patterns remain consistent across domains when the semantic content is preserved (\Cref{APPENDIX:SUBSECTION_3_4}).}
    \label{Fig-Rebuttal-A}
    \end{center}
    \vskip -0.2in
\end{figure}

To further evaluate whether the model’s internal mechanisms generalize beyond natural images, we analyze its behavior under more challenging domain shifts. While \cref{Fig-2A}E demonstrates that semantically similar natural images elicit consistent attention head usage patterns, all tested inputs were restricted to natural images. Here, we extend the analysis to the DomainNet dataset~\citep{peng2019moment}, which contains images from diverse visual domains, including real, sketch, and clipart styles. This setting enables a more rigorous test of the model’s robustness and invariance to domain-specific variations. We randomly sample 300 images across the three domains (see \Cref{Fig-Rebuttal-A}D for examples), and apply our head attribution method to examine whether semantically equivalent images still activate similar attention head patterns.

First, we confirm that the proposed head attribution method maintains high faithfulness and completeness compared to the Attention and Causal baselines (\Cref{Fig-Rebuttal-A}A), in line with the results in Section 4.2. \Cref{Fig-Rebuttal-A}B and C replicate the analyses of \cref{Fig-2A}C and D, showing that head attribution patterns remain consistent across different domains. This consistency suggests that the model’s attention heads are not tailored to specific visual domains but rather capture domain-invariant semantic structures. Finally, \Cref{Fig-Rebuttal-A}E illustrates that attention head usage patterns cluster similarly across styles, as observed previously in natural image domains. This further supports the conclusion that the model generalizes well across domains and that its internal mechanisms reflect semantic content rather than superficial style.

Overall, these results demonstrate that our finding extends beyond natural images and remains effective across diverse visual domains, validating its generalizability in broader settings.

\subsection{Attribution Coefficients of Individual Attention Heads}
\label{APPENDIX:SUBSECTION_3_5}

\begin{figure}
    \begin{center}
    \centerline{\includegraphics[width=\columnwidth]{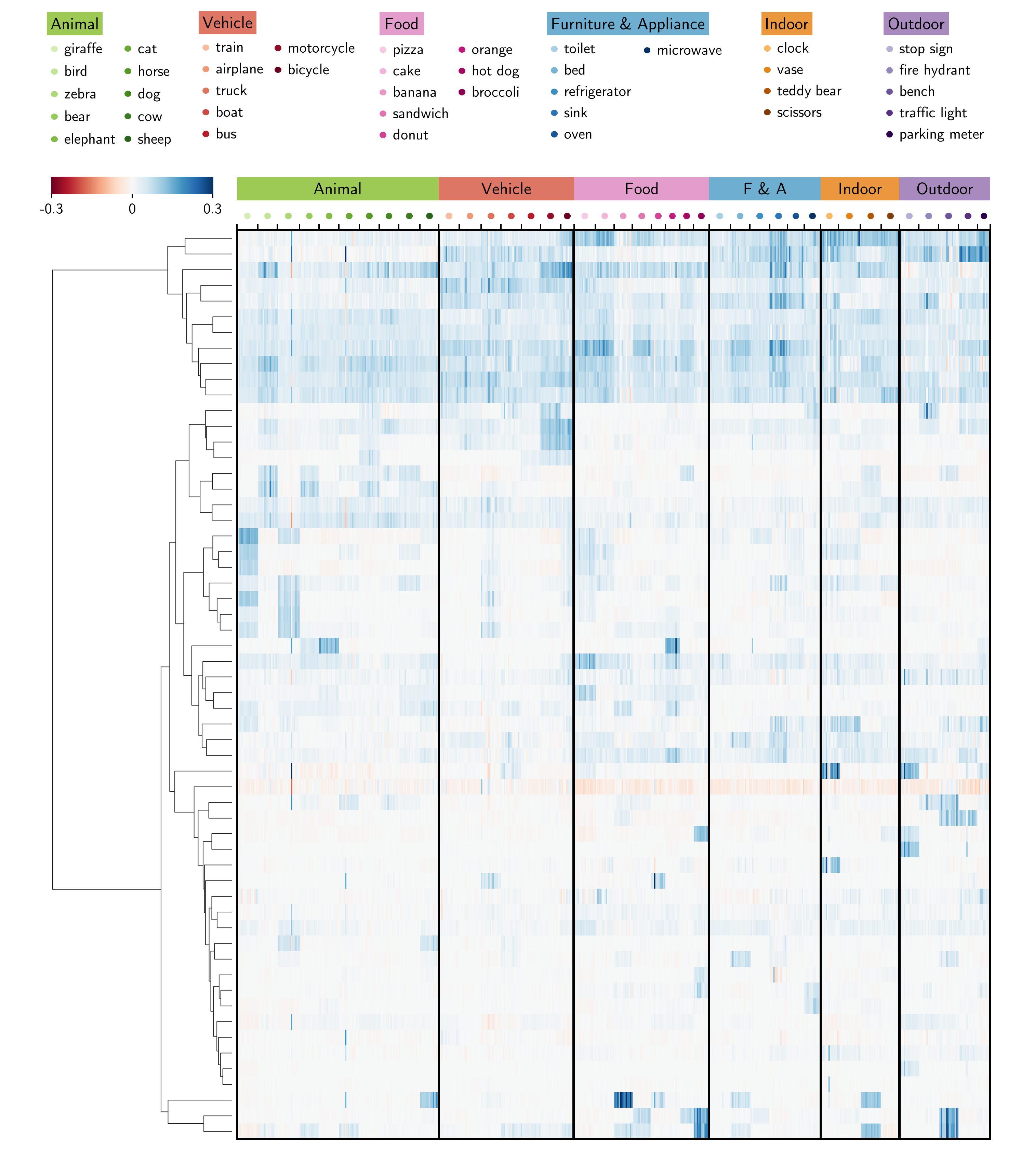}}
    \vskip -0.1in
    \caption{Visualization of the head vectors $\boldsymbol{h}_n$ of attention heads with $\theta_n > 0.1$ in at least one sample in LLaVA-1.5-7B. The head vectors are hierarchically clustered. The x-axis represents each sample, and the y-axis represents each attention head. See \cref{APPENDIX:SUBSECTION_3_5} for the interpretation.}
    \label{Fig-XC2}
    \end{center}
    \vskip -0.2in
\end{figure}

To gain a deeper understanding of individual attention heads, we concatenate the attribution coefficients $\theta_n$ of each head $n$ across all samples used in \cref{Fig-2A}E and define this as a head vector $\boldsymbol{h}_n$. We then visualize the head vectors of attention heads with $\theta_n > 0.1$ in at least one sample in \cref{Fig-XC2}. To identify patterns among attention heads, we apply hierarchical clustering to the head vectors, grouping similar attention heads together. The results are interpreted as follows.

First, the head vectors form two major clusters. The first cluster consists of attention heads that are consistently utilized across samples (sample-agnostic heads), while the second cluster comprises heads that are used in a sample-specific manner (sample-specific heads). This finding aligns with the observation in \cref{Fig-2A}C, indicating that LVLMs leverage both sample-agnostic and sample-specific attention heads to process image-to-text information flow.

Second, the head vector patterns remain consistent within the same object category or super-category. In other words, different images containing the same object tend to activate the same attention heads. This observation further supports the results in \cref{Fig-2A}E, demonstrating that LVLMs utilize attention heads based on the semantics of the input image.

\section{Additional Results for Other Models}
\label{APPENDIX:ADDITIONAL}
In this section, we provide additional results for other models.

In \cref{SUBSECTION_4_1}, we fit a linear regression model to estimate the final logit based on which attention heads remain intact. We report the quantitative results of the regression model, including the explained variance ($R^2$) and the Pearson correlation coefficient ($\rho$), in \Cref{Table-XD1}. High $R^2$ and $\rho$ values indicate that the contribution of attention heads to the final logit is well captured by the linear regression model.

\begin{figure}[ht]
    \vskip 0in
    \begin{center}
    \centerline{\includegraphics[width=1\columnwidth]{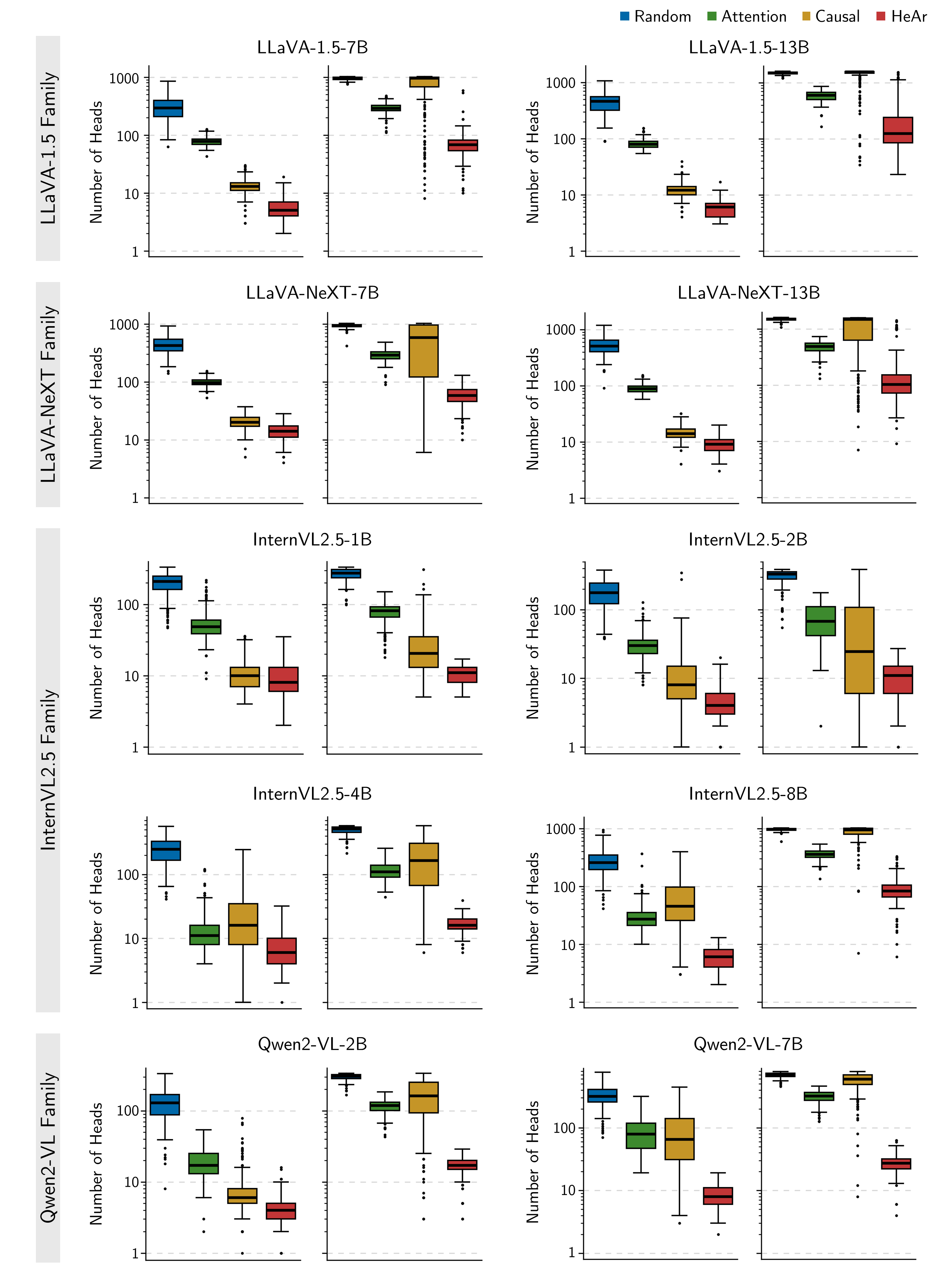}}
    \caption{The minimum number of attention heads for faithfulness $> 0.8$ (Left) and completeness $< 0.2$ (Right) across all models. The head attribution method (HeAr) needs fewer heads to achieve the same level of faithfulness and completeness compared to the other methods. Related to \cref{Fig-2A}A.}
    \label{Fig-XD1}
    \end{center}
    \vskip -0.2in
\end{figure}

In \cref{SUBSECTION_4_2}, we validate the head attribution method by comparing the number of heads required to achieve the same level of faithfulness and completeness across three other methods: Random, Attention, and Causal. We report the full results in \cref{Fig-XD1}. The head attribution method consistently requires fewer heads to achieve the same level of faithfulness and completeness, demonstrating its effectiveness in measuring the importance of attention heads.

\begin{figure}[ht]
    \vskip 0in
    \begin{center}
    \centerline{\includegraphics[width=\columnwidth]{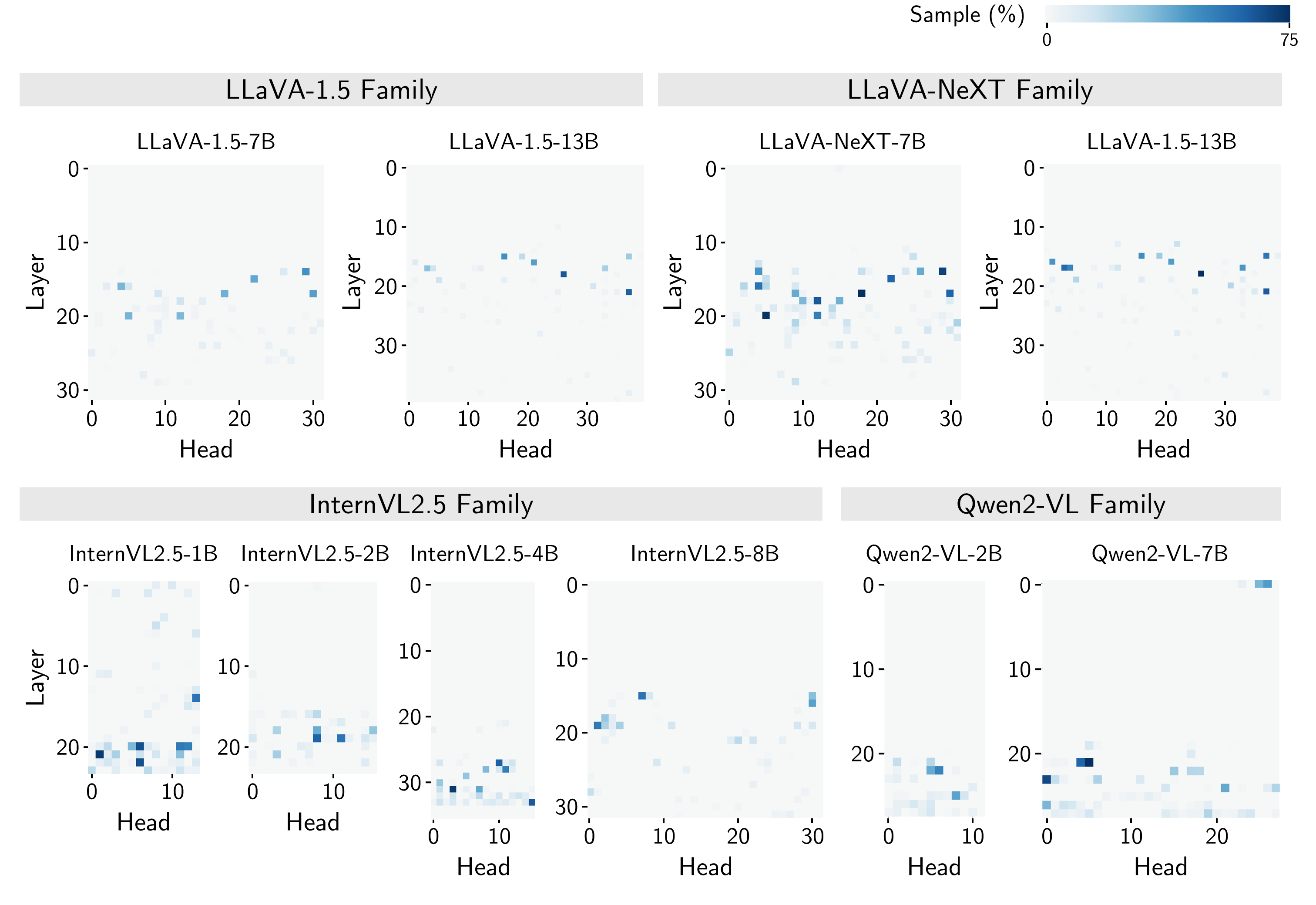}}
    \caption{The portion of samples requiring a given head for faithfulness $> 0.8$ across all models. The important attention heads are located in the mid-to-late layers. Related to \cref{Fig-2A}C.}
    \label{Fig-XD2}
    \end{center}
\end{figure}

\begin{figure}[ht]
    \vskip 0in
    \begin{center}
    \centerline{\includegraphics[width=\columnwidth]{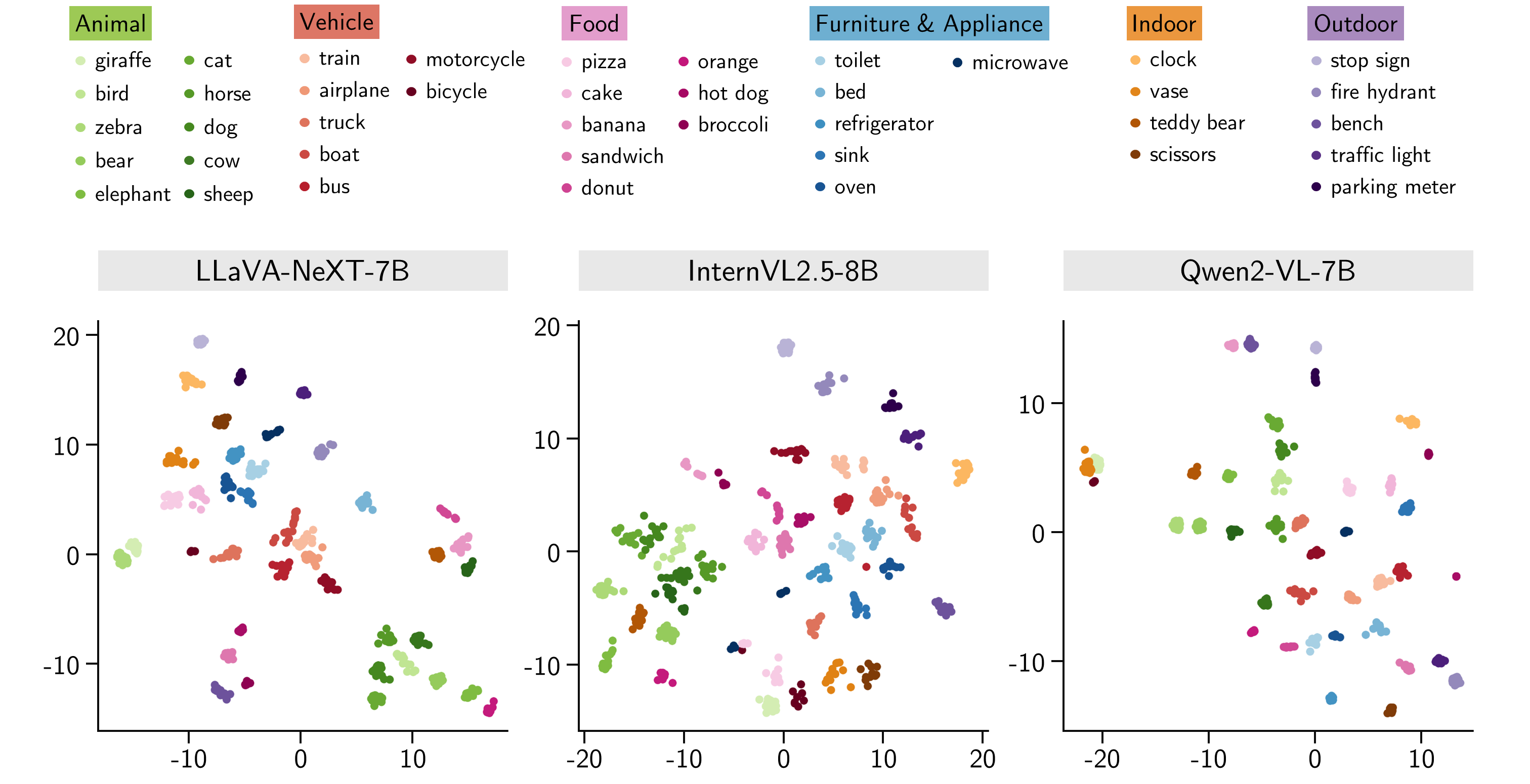}}
    \caption{t-SNE visualization of attribution coefficients $\boldsymbol{\theta}$ in LLaVA-NeXT-7B, InternVL2.5-8B, and Qwen2-VL-7B. Objects with similar semantic meanings are typically clustered together, indicating that LVLMs systematically utilize attention heads to process the input image based on their semantic contents. Related to \cref{Fig-2A}E.}
    \label{Fig-XD4}
    \end{center}
\end{figure}

\begin{figure}[ht]
    \vskip 0.2in
    \begin{center}
    \centerline{\includegraphics[width=\columnwidth]{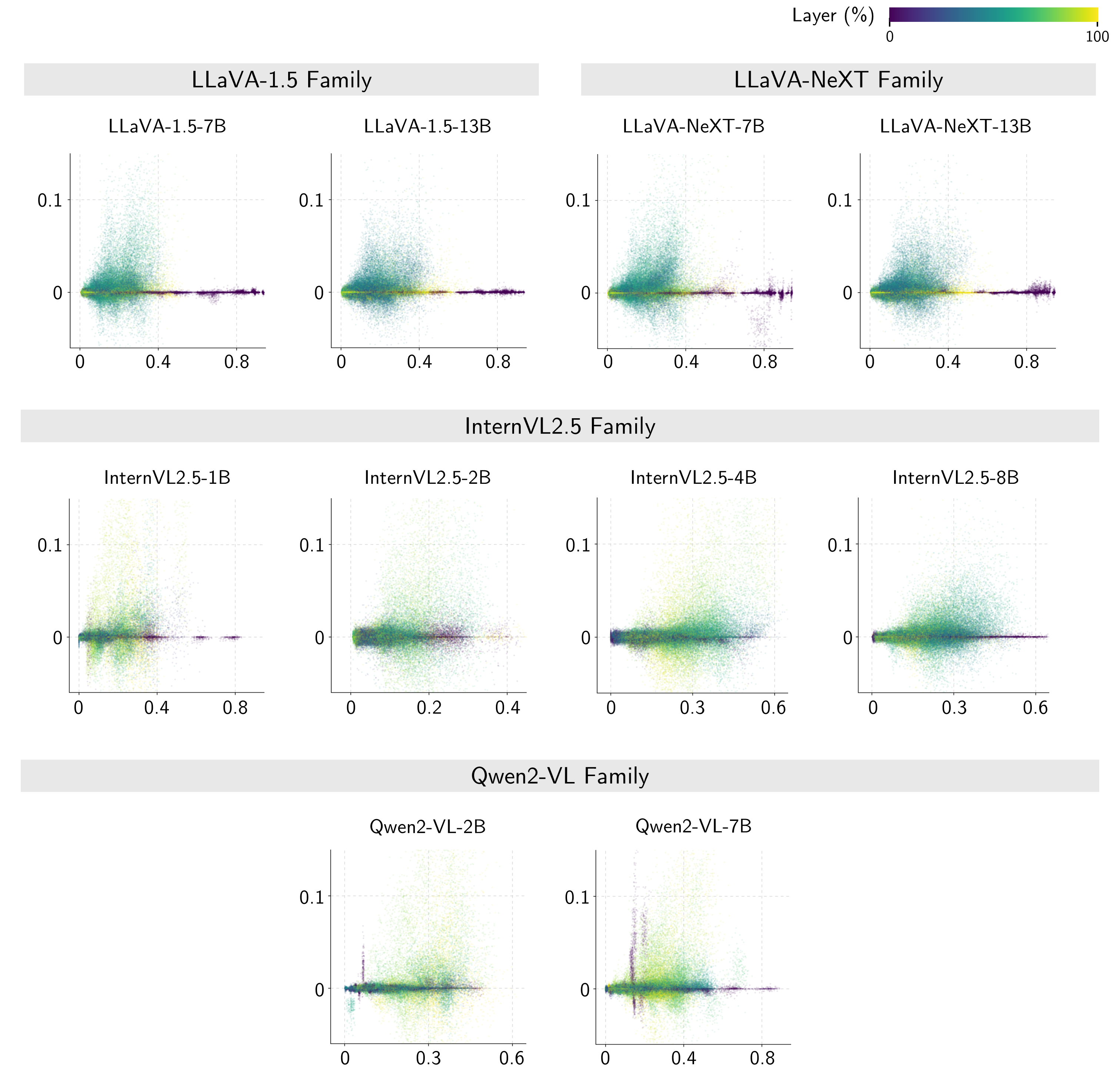}}
    \caption{Scatter plot showing the relationship between attribution coefficients $\boldsymbol{\theta}$ and image attention weights across all models. The x-axis represents image attention weights, and the y-axis represents attribution coefficients. The color of each point represents the relative layer depth. No significant correlation is observed between the two, suggesting that image attention weights are not reliable for explaining the importance of attention heads. Related to \cref{Fig-2A}D.}
    \label{Fig-XD3}
    \end{center}
    \vskip -0.2in
\end{figure}

In \cref{SUBSECTION_4_3}, we analyze the head attribution results from three perspectives: the distribution of important attention heads (\cref{Fig-2A}C), their relation to image attention weights (\cref{Fig-2A}D), and systematic patterns in attention heads (\cref{Fig-2A}E). \cref{Fig-XD2} shows the distribution of important attention heads across all models. Generally, important heads are concentrated in mid-to-late layers. Interestingly, LLaVA-1.5 and LLaVA-NeXT, which share the same backbone large language model, exhibit similar distributions of important attention heads. This result suggests that, to some extent, the role of attention heads is predetermined by the backbone large language model. We leave further investigation of this phenomenon for future work. \cref{Fig-XD3} visualizes the relationship between attribution coefficients $\boldsymbol{\theta}$ and image attention weights. No clear correlation is observed between the two, supporting the claim that image attention weights are not reliable indicators of attention head importance. \cref{Fig-XD4} presents the t-SNE visualization of the attribution coefficients $\boldsymbol{\theta}$ in LLaVA-NeXT-7B, InternVL2.5-8B, and Qwen2-VL-7B. We select representative models from each family due to the computational cost of collecting additional samples. The t-SNE visualization reveals that LVLMs systematically utilize attention heads in response to the semantic content of the input, as samples belonging to the same super-category are typically clustered together.

\begin{figure}[ht]
    \vskip 0.2in
    \begin{center}
    \centerline{\includegraphics[width=\columnwidth]{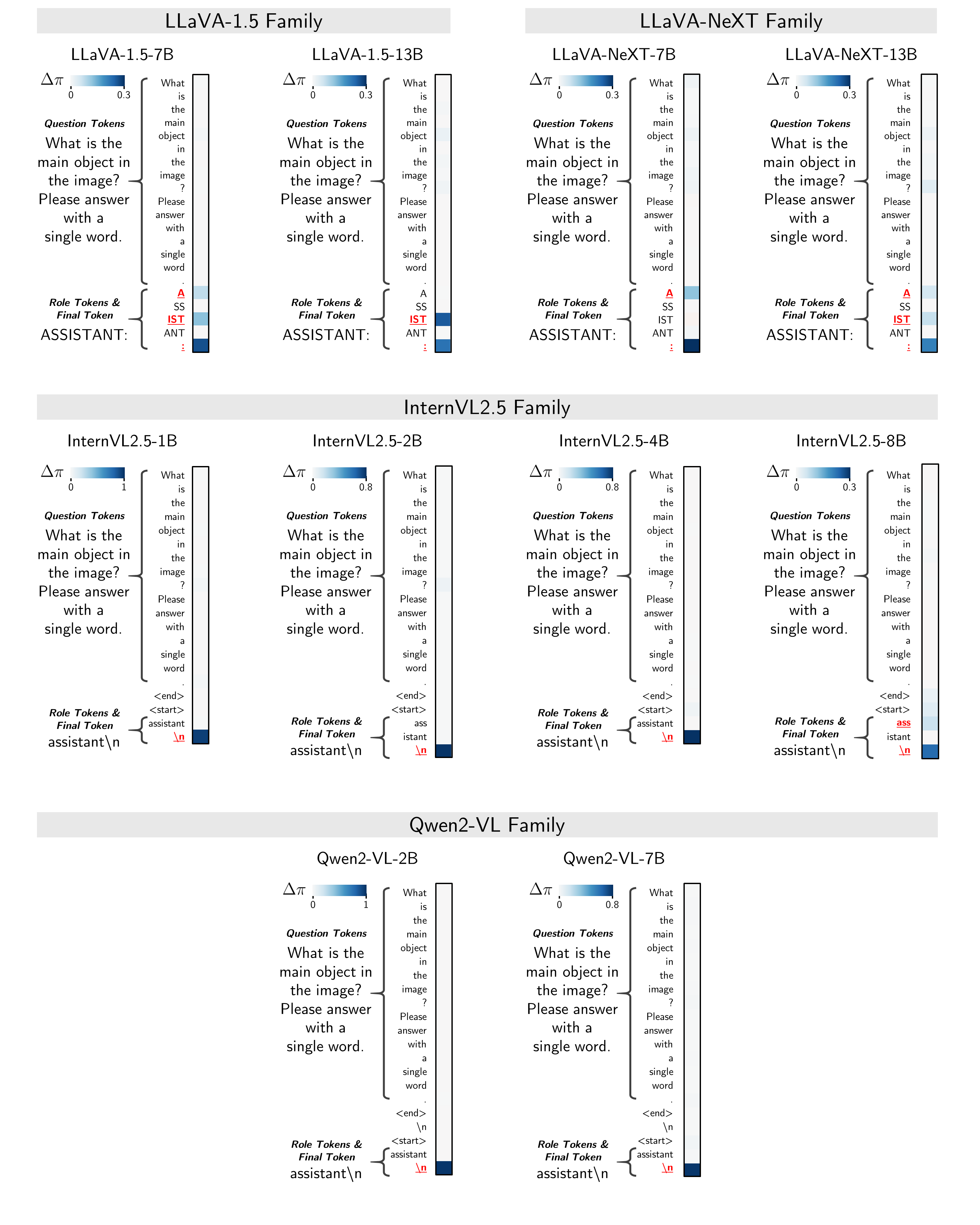}}
    \caption{Logit difference relative to $\pi (\boldsymbol{x}^\ast)$ when blocking each text token across all models. The text tokens with $\Delta \pi_i > 0.05$ are highlighted in red, which are consistently the role tokens and the final token. Related to \cref{Fig-4A}A.}
    \label{Fig-XD5}
    \end{center}
\end{figure}

In \cref{SUBSECTION_5_1}, we identify the text tokens that receive image information by measuring the causal effect of blocking image information flow to each text token. As shown in \cref{Fig-XD5}, these tokens are primarily role tokens and the final token in the text. Furthermore, we validate these findings by measuring the logit when retaining only the role tokens and the final token. More specifically, we retain tokens with $\Delta \pi_i > 0.05$ (see \cref{APPENDIX:SUBSECTION_1_4} for the definition), which are highlighted in red in \cref{Fig-XD5}. As shown in \Cref{Table-XD2}, the logit remains well preserved, indicating that receiving image information in the role tokens and the final token is essential for the model to predict the correct answer.

\begin{figure}[ht]
    \vskip 0.2in
    \begin{center}
    \centerline{\includegraphics[width=\columnwidth]{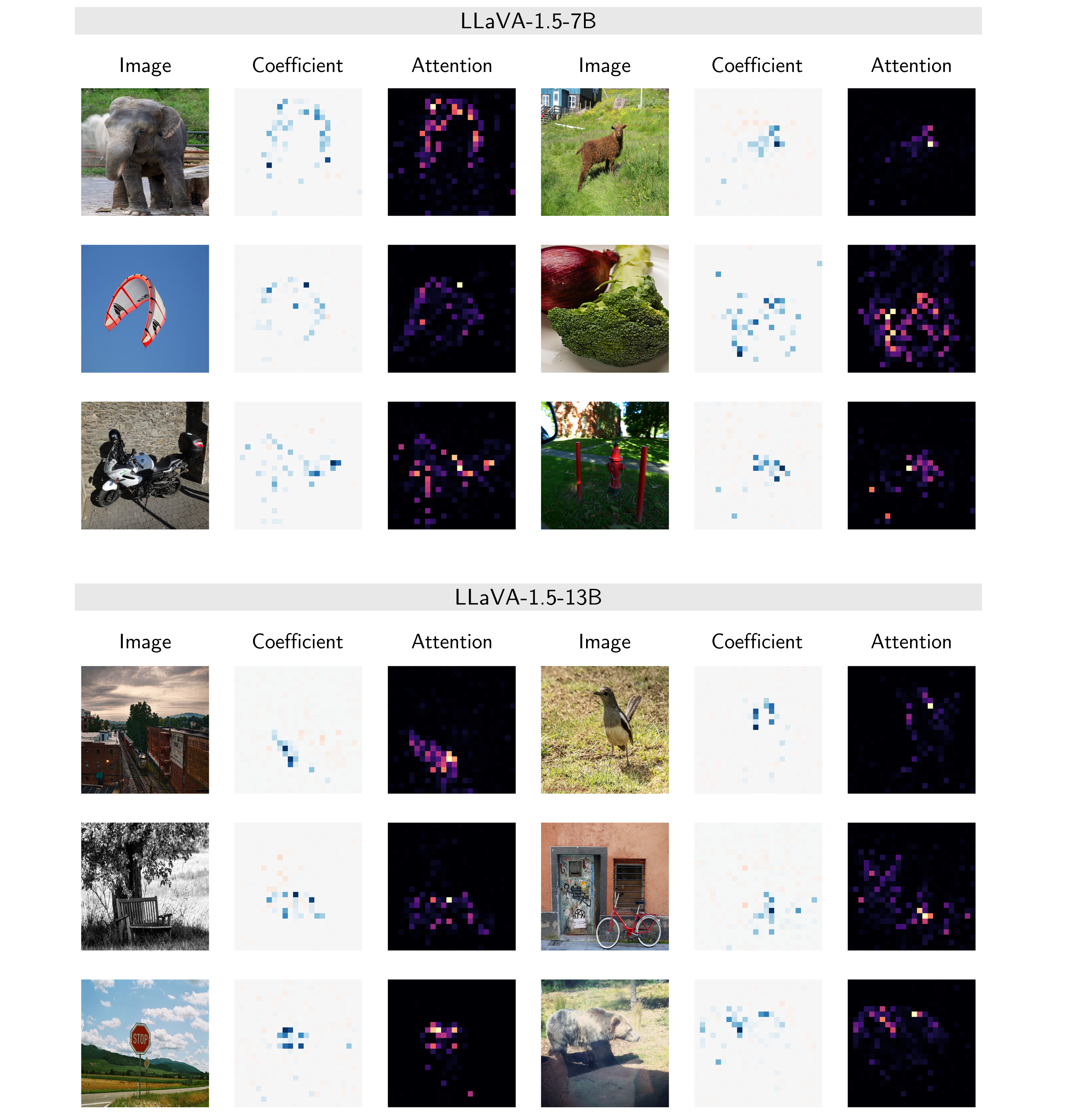}}
    \caption{More qualitative results of image, attribution coefficients $\boldsymbol{\theta}$, and image attention weights for LLaVA-1.5-7B and LLaVA-1.5-13B. Related to \cref{Fig-4A}B.}
    \label{Fig-XD6}
    \end{center}
    \vskip -0.2in
\end{figure}

\begin{figure}[ht]
    \vskip 0.2in
    \begin{center}
    \centerline{\includegraphics[width=\columnwidth]{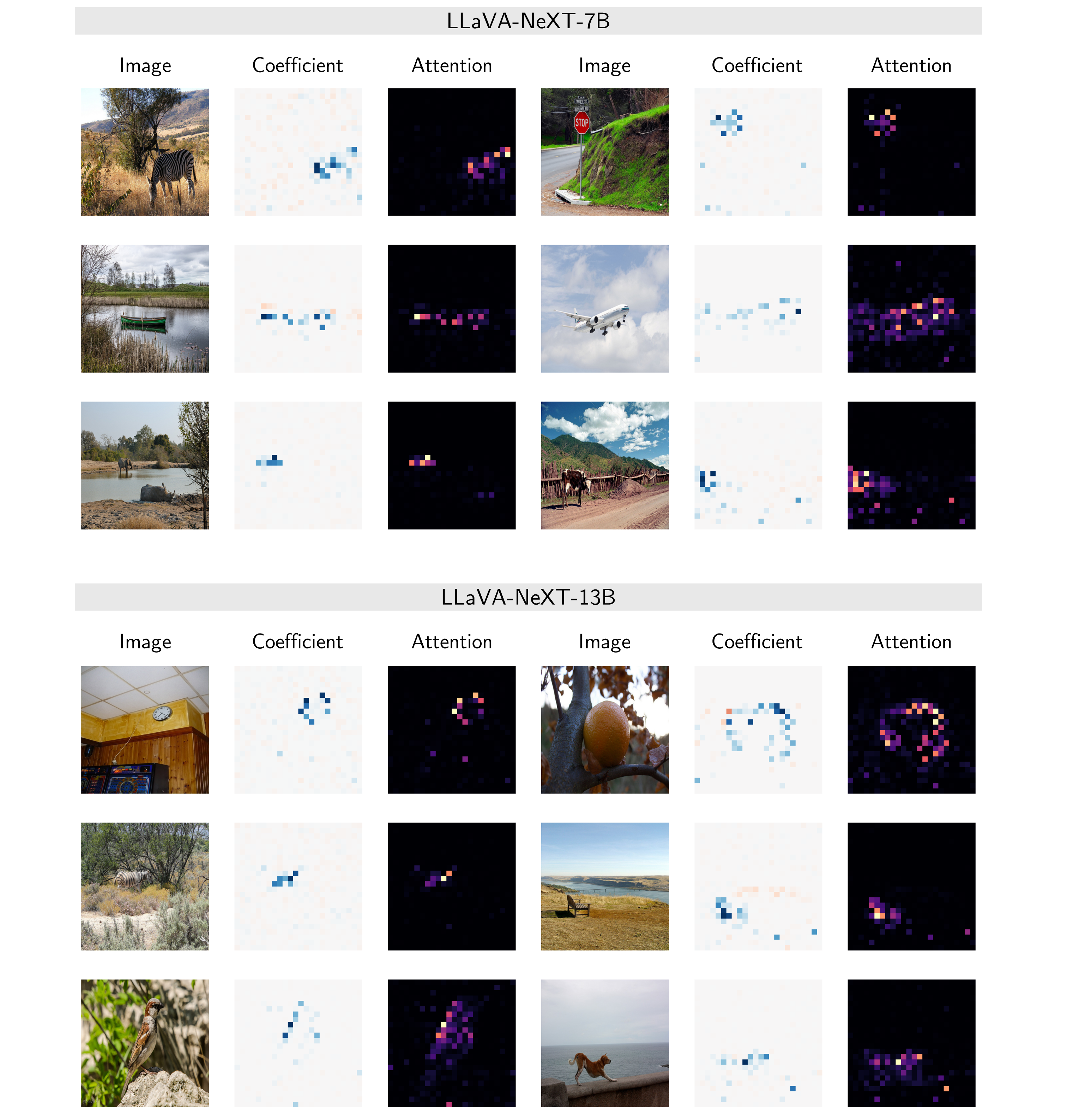}}
    \caption{More qualitative results of image, attribution coefficients $\boldsymbol{\theta}$, and image attention weights for LLaVA-NeXT-7B and LLaVA-NeXT-13B. Related to \cref{Fig-4A}B.}
    \label{Fig-XD7}
    \end{center}
    \vskip -0.2in
\end{figure}

In \cref{SUBSECTION_5_2}, we qualitatively visualize the attribution coefficients $\boldsymbol{\theta}$ and image attention weights (see \cref{APPENDIX:SUBSECTION_1_4} for technical details). Additional qualitative results are presented in \cref{Fig-XD6} and \cref{Fig-XD7}. The results indicate that only a subset of object tokens and a few background tokens contribute to the final logit, while the remaining tokens have minimal impact. Notably, the number of important tokens is smaller than the number of tokens with high image attention weights, suggesting that further token reduction is possible without significantly affecting the model's performance.

\newpage

\begin{table}[h]
\caption{Quantitative results of head attribution for all models. The table presents the explained variance ($R^2$) and the Pearson correlation coefficient ($\rho$) between $\pi(\boldsymbol{x})$ and $\hat{\pi}(\boldsymbol{x})$. These results indicate that the contribution of attention heads to the final logit is well captured by the linear regression model. Related to \cref{SUBSECTION_4_1}.}
\label{Table-XD1}
\vskip 0.15in
\begin{center}
\begin{tabular}{lcc}
\toprule
Model & $R^2$ & $\rho$ \\
\midrule
LLaVA-1.5-7B & $0.82 \pm 0.05$ & $0.91 \pm 0.03$ \\
LLaVA-1.5-13B & $0.79 \pm 0.06$ & $0.89 \pm 0.03$ \\
\midrule
LLaVA-NeXT-7B & $0.83 \pm 0.05$ & $0.91 \pm 0.03$ \\
LLaVA-NeXT-13B & $0.82 \pm 0.04$ & $0.91 \pm 0.02$ \\
\midrule
InternVL2.5-1B & $0.92 \pm 0.03$ & $0.96 \pm 0.02$ \\
InternVL2.5-2B & $0.79 \pm 0.05$ & $0.89 \pm 0.03$ \\
InternVL2.5-4B & $0.79 \pm 0.06$ & $0.89 \pm 0.03$ \\
InternVL2.5-8B & $0.77 \pm 0.06$ & $0.88 \pm 0.03$ \\
\midrule
Qwen2-VL-2B & $0.90 \pm 0.04$ & $0.95 \pm 0.02$ \\
Qwen2-VL-7B & $0.82 \pm 0.05$ & $0.91 \pm 0.03$ \\
\bottomrule
\end{tabular}
\end{center}
\vskip -0.1in
\end{table}
\begin{table}[h]
\caption{The logit relative to $\pi (\boldsymbol{x}^\ast)$ when retaining only the important text tokens ($\Delta \pi_i > 0.05$, highlighted in red in \cref{Fig-XD5}). These results suggest that the role tokens and the final token are essential for the model to predict the correct answer. Related to \cref{SUBSECTION_5_1}.}
\label{Table-XD2}
\vskip 0.15in
\begin{center}
\begin{tabular}{lc}
\toprule
Model & Logit relative to $\pi{(\boldsymbol{x}^\ast)}$ \\
\midrule
LLaVA-1.5-7B & $0.98 \pm 0.02$ \\
LLaVA-1.5-13B & $0.92 \pm 0.04$ \\
\midrule
LLaVA-NeXT-7B & $0.95 \pm 0.02$ \\
LLaVA-NeXT-13B & $0.89 \pm 0.05$ \\
\midrule
InternVL2.5-1B & $0.94 \pm 0.15$ \\
InternVL2.5-2B & $0.93 \pm 0.07$ \\
InternVL2.5-4B & $0.93 \pm 0.04$ \\
InternVL2.5-8B & $0.82 \pm 0.14$ \\
\midrule
Qwen2-VL-2B & $0.99 \pm 0.01$ \\
Qwen2-VL-7B & $0.82 \pm 0.10$ \\
\bottomrule
\end{tabular}
\end{center}
\vskip -0.1in
\end{table}







\end{document}